\documentclass[10pt,twocolumn,letterpaper]{article}

\usepackage{iccv}
\usepackage{times}
\usepackage{epsfig}
\usepackage{graphicx}
\usepackage{amsmath}
\usepackage{amssymb}
\usepackage[accsupp]{axessibility}  

\usepackage{subcaption}
\usepackage{booktabs}
\usepackage{multirow}
\usepackage{makecell}
\usepackage[ruled]{algorithm2e}
\usepackage[pagebackref=true,breaklinks=true,letterpaper=true,colorlinks,bookmarks=false]{hyperref}

\def\DD{\mathcal{D}}

\DeclareMathOperator*{\argmax}{argmax}

\newtheorem{theorem}{Theorem}
\newtheorem{definition}{Definition}

\usepackage[capitalize]{cleveref}
\crefname{section}{Sec.}{Secs.}
\Crefname{section}{Section}{Sections}
\Crefname{table}{Table}{Tables}
\crefname{table}{Tab.}{Tabs.}
\iccvfinalcopy 

\def\iccvPaperID{6082} 

\ificcvfinal\fi

\begin{document}
\title{Label-Noise Learning with Intrinsically Long-Tailed Data}
\author{
Yang Lu\textsuperscript{\rm 1,2}\thanks{Yang Lu is the corresponding author: luyang@xmu.edu.cn} \quad  Yiliang Zhang\textsuperscript{\rm 1,2} \quad Bo Han\textsuperscript{\rm 3} \quad Yiu-ming Cheung\textsuperscript{\rm 3} \quad  Hanzi Wang\textsuperscript{\rm 1,2}\\
\small{\textsuperscript{\rm 1}Fujian Key Laboratory of Sensing and Computing for Smart City, School of Informatics, Xiamen University, Xiamen, China}\\
\small{\textsuperscript{\rm 2}Key Laboratory of Multimedia Trusted Perception and Efficient Computing,}\\
\small{Ministry of Education of China, Xiamen University, Xiamen, China}\\
\small{\textsuperscript{\rm 3}Department of Computer Science, Hong Kong Baptist University, Hong Kong, China}\\
\tt\footnotesize{luyang@xmu.edu.cn   ylzhangcs@hotmail.com    \{bhanml, ymc\}@comp.hkbu.edu.hk    hanzi.wang@xmu.edu.cn}
}
\maketitle
\ificcvfinal\thispagestyle{empty}\fi

\begin{abstract}
Label noise is one of the key factors that lead to the poor generalization of deep learning models. Existing label-noise learning methods usually assume that the ground-truth classes of the training data are balanced.
However, the real-world data is often imbalanced, leading to the inconsistency between observed and intrinsic class distribution with label noises.
In this case, it is hard to distinguish clean samples from noisy samples on the intrinsic tail classes with the unknown intrinsic class distribution.
In this paper, we propose a learning framework for label-noise learning with intrinsically long-tailed data.
Specifically, we propose two-stage bi-dimensional sample selection (TABASCO) to better separate clean samples from noisy samples, especially for the tail classes.
TABASCO consists of two new separation metrics that complement each other to compensate for the limitation of using a single metric in sample separation.
Extensive experiments on benchmarks  demonstrate the effectiveness of our method.
Our code is available at \color{magenta}\url{https://github.com/Wakings/TABASCO}\color{black}.
\end{abstract}

\vspace{-12px}
\section{Introduction}

Under the support of a large amount of high-quality labeled data, deep neural networks have achieved great success in various fields \cite{DBLP:conf/nips/KrizhevskySH12,DBLP:conf/nips/RenHGS15,DBLP:conf/naacl/DevlinCLT19}. However, it is expensive and difficult to obtain a large amount of high-quality labeled data in many practical applications.
Instead, the commonly used large-scale training data is usually obtained from the Internet or crowdsourcing platforms like Amazon Mechanical Turk, which is unreliable and may be mislabeled \cite{DBLP:conf/cvpr/XiaoXYHW15,DBLP:conf/icml/SongK019}.
The models trained on this kind of unreliable data, called noisy-labeled data, often produce poor generalization performance because deep neural networks tend to overfit noisy samples due to their large model capacity. 
In the literature, there are some works to obtain a robust model trained on noisy-labeled data \cite{DBLP:journals/corr/abs-2011-04406,DBLP:journals/corr/abs-2007-08199}. 
Among them, the most straightforward and effective way is to differentiate between clean and noisy samples based on their differences in specific metrics, such as training loss \cite{DBLP:conf/nips/HanYYNXHTS18,DBLP:conf/iclr/LiSH20,Karim_2022_CVPR}. 

\begin{figure}[!t]
\begin{subfigure}[b]{0.47\columnwidth}
    \hspace{-3px}
    \includegraphics[width=\textwidth]{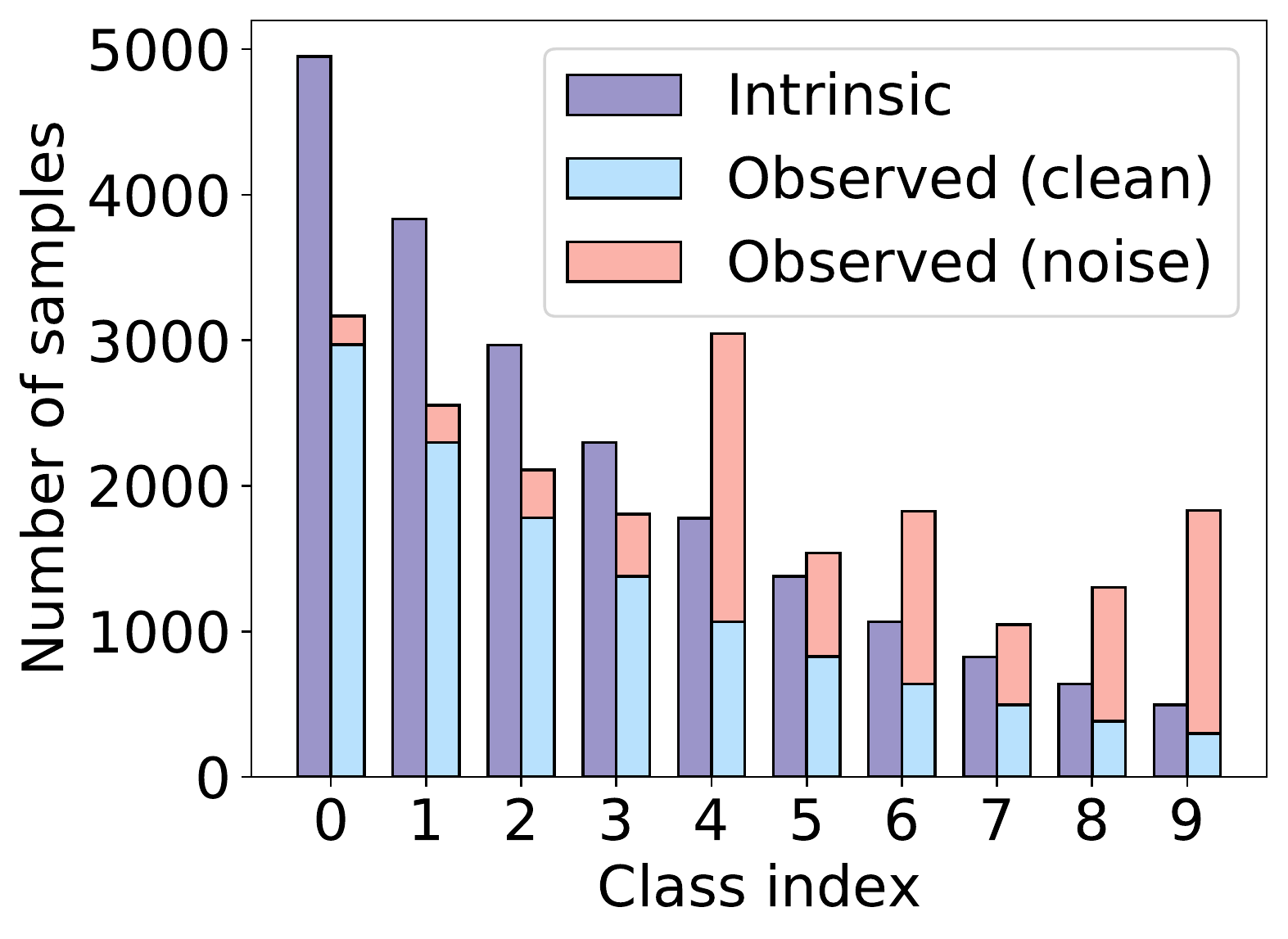}
    \subcaption{Class distribution}
\end{subfigure}
\hspace{-1px}
\begin{subfigure}[b]{0.47\columnwidth}
    \includegraphics[width=\textwidth]{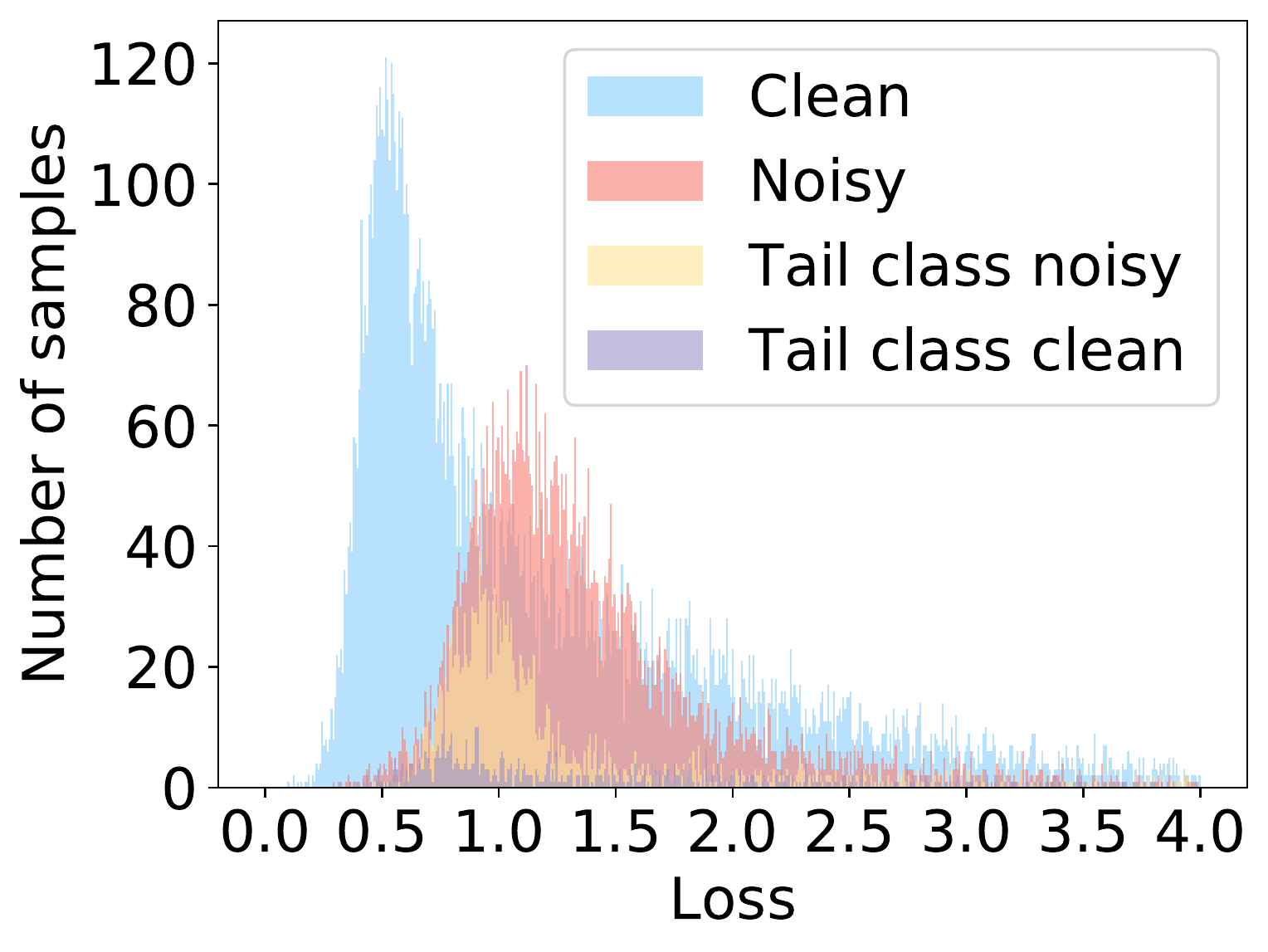}
    \subcaption{Loss distribution}
\end{subfigure}
\vspace{-5px}
\caption{(a) An example of observed class distribution with noisy labels with long-tailed intrinsic class distribution. (b) The training loss of each sample under noisy-labeled and long-tailed data.}
\label{motivation}
\vspace{-15px}
\end{figure}

These label-noise learning methods generally assume that the intrinsic class distribution of the training data is balanced, where each class has almost the same number of samples in terms of their unknown ground-truth labels.
However, the data in real-world applications are often imbalanced, e.g., the LVIS dataset \cite{gupta2019lvis} and the iNaturalist dataset \cite{DBLP:conf/cvpr/HornASCSSAPB18}.
The class imbalance usually exhibits in the form of a long-tail  distribution, where a small portion of classes possess a large number of samples, and the other classes possess a small number of samples only \cite{REED200115,DBLP:journals/corr/abs-1709-01450}.
In this case, the model training tends to the head classes and ignores the tail classes \cite{DBLP:conf/aaai/ZhangWZW21,DBLP:conf/aaai/ZhuN0Z22}.
When both noisy labels and long-tail distribution exist, training a robust model is even more challenging. 
There are two key challenges in this scenario. 
(1) \textbf{\textit{Distribution inconsistency}}:
The observed and intrinsic distributions are likely inconsistent due to noise labels, making the model more difficult to discover and focus on the intrinsic tail classes.
As illustrated in \cref{motivation}(a), the intrinsic class distribution of the dataset is long-tailed, while the existence of noisy labels makes the distribution more balanced.
The intrinsic tail classes, e.g., classes 6 and 9, are occupied by a large number of noisy data, making them no longer tail classes by observation.
(2) \textbf{\textit{Tail inseparability}}:
Even if the tail class is identified, it is more difficult than ever to distinguish between clean and noisy samples in the tail class because clean samples are overwhelmed by noises that make their values of the separation metric highly similar.
As illustrated in \cref{motivation}(b), the training loss of clean and noisy samples in the tail class are generally inseparable compared with the ones in the other classes.
Several preliminary works have studied the joint problem of  label noise and long-tail distribution \cite{jiang2022delving,DBLP:journals/corr/abs-2108-11569,DBLP:journals/corr/abs-2108-11096,DBLP:conf/iclr/CaoCLAGM21}.
These methods implicitly reduce the complexity of the problem by assuming a similar noise rate for each class.
However, this assumption is too strong to apply because noisy samples from the head classes may be the majority, resulting in a higher noise rate in the tail class than in other classes.

In this paper, we propose a Two-stAge Bi-dimensionAl Sample seleCtiOn (TABASCO) strategy to address the problem of label-noise learning with intrinsically long-tailed data.
In the first stage, we propose to use two new separation metrics for sample separation, i.e., weighted Jensen-Shannon divergence (WJSD) and adaptive centroid distance (ACD), which work corporately to separate clean samples from noisy samples in the tail classes.
The proposed metrics are complementary, where WJSD separates the samples from the output perspective while ACD does that from the feature perspective.
In the second stage, we determine the separation dimension with better separability for each class and perform sample selection.
In order to evaluate the method uniformly and effectively, we introduce two benchmarks with real-world noise and intrinsically long-tailed distribution. 
The main contributions of our work can be summarized as follows:
\begin{itemize}
    \setlength{\itemsep}{2pt}
    \setlength{\parsep}{2pt}
    \setlength{\parskip}{2pt}
    \item We present a more general problem of label-noise learning with intrinsically long-tailed data. The key challenges in this problem are distribution inconsistency and tail inseparability.
    \item We propose an effective solution called TABASCO. With the help of two new separation metrics, it is able to effectively identify and select clean samples of the intrinsic tail class.
    \item We introduce two benchmarks with real-world noise and intrinsically long-tailed distribution. Extensive experiments on them show the effectiveness of our method and the limitations of existing methods.
\end{itemize}
\section{Related Work}
\label{sec:formatting}

\subsection{Label-Noise Learning}
A straightforward strategy to deal with noisy data is to reduce the proportion of noise in training samples by separating noisy samples from clean samples. 
Methods such as co-teaching \cite{DBLP:conf/nips/HanYYNXHTS18} and DivideMix \cite{DBLP:conf/iclr/LiSH20} adopt the small loss trick, while Jo-SRC \cite{DBLP:conf/cvpr/YaoSZS00T21} and UNICON \cite{Karim_2022_CVPR} use Jensen-Shannon divergence instead for sample selection. 
In contrast to methods that separate at the label level, some methods \cite{Li_2022_CVPR,DBLP:conf/cvpr/OrtegoAAOM21} attempt to separate samples at the feature space.
There are also some methods to avoid over-fitting the model to noisy data by imposing regularization constraints on model parameters \cite{DBLP:conf/iclr/XiaL00WGC21,DBLP:conf/icml/00030YYXTS20} or labels \cite{DBLP:conf/iclr/PereyraTCKH17,DBLP:conf/iclr/ZhangCDL18,DBLP:conf/icml/LukasikBMK20}. 
Other methods mitigate the influence of noisy data by adjusting the loss functions, such as backward and forward loss correction \cite{DBLP:conf/cvpr/PatriniRMNQ17}, gold loss correction \cite{DBLP:conf/nips/HendrycksMWG18}, MW-Net \cite{DBLP:conf/nips/ShuXY0ZXM19} and Dual-T \cite{DBLP:conf/nips/YaoL0GD0S20}. 

\subsection{Long-tail Learning}
Re-balancing the data for long-tail distributions is a classical strategy to solve the problem of long-tail learning, such as re-sampling \cite{DBLP:journals/jair/ChawlaBHK02,DBLP:journals/tsmc/LiuWZ09,DBLP:journals/ci/EstabrooksJJ04,DBLP:conf/icic/HanWM05} and data augmentation \cite{DBLP:conf/cvpr/ZhongC0J21,DBLP:conf/iccv/ZangHL21,DBLP:conf/eccv/ChouCPWJ20}. 
In addition, there are methods to improve the model generalization by introducing long-tail robust loss functions \cite{DBLP:conf/cvpr/CuiJLSB19,DBLP:conf/nips/CaoWGAM19,DBLP:conf/cvpr/TanWLLOYY20,DBLP:conf/nips/RenYSMZYL20}. Methods such as FTL \cite{DBLP:conf/cvpr/00010S0C19}, RIDE \cite{DBLP:conf/iclr/WangLM0Y21} and DiVE \cite{DBLP:conf/iccv/HeWW21} try to solve the problem by using the idea of transfer learning. 
Recently, approaches based on decoupling \cite{DBLP:conf/iclr/KangXRYGFK20,DBLP:conf/cvpr/ZhongC0J21,DBLP:conf/cvpr/ZhangLY0S21} split end-to-end learning into feature learning and classifier retraining such that the obtained feature extractor is less affected by the long-tail distribution. 
In contrast to the above methods with the supervised learning paradigm, CReST \cite{DBLP:conf/cvpr/WeiSMYY21}, ABC \cite{DBLP:conf/nips/LeeSK21} and DARP \cite{DBLP:conf/nips/KimHPYHS20} attempt to solve the long-tail problem in the manner of semi-supervised learning.

\subsection{Label-Noise Learning on Long-tailed Data}
Research on this joint problem has just been explored. 
CNLCU \cite{xia2022sample} relaxes the constraint of the small loss trick by regarding a portion of large loss samples as clean samples to reduce the probability of misclassifying clean samples to noisy samples in the tail class. 
RoLT \cite{DBLP:journals/corr/abs-2108-11569} uses the distance from the samples to the centroid of the current class instead of the training loss for sample selection. 
HAR \cite{DBLP:conf/iclr/CaoCLAGM21} uses an adaptive approach to regularize noise and tail class samples. 
Karthik et al. \cite{DBLP:journals/corr/abs-2108-11096} uses the idea of decoupling to fine-tune the loss function for better robustness after feature learning by the self-supervised method.
ULC \cite{DBLP:conf/aaai/HuangBZBW22} introduces uncertainty to enhance the separation ability of noise samples.
H2E \cite{yi2022identifying} reduces hard noises to easy ones by learning a classifier as noise identifier invariant to the class and context distributional changes.
\begin{figure*}[!t]
\includegraphics[width=\textwidth]{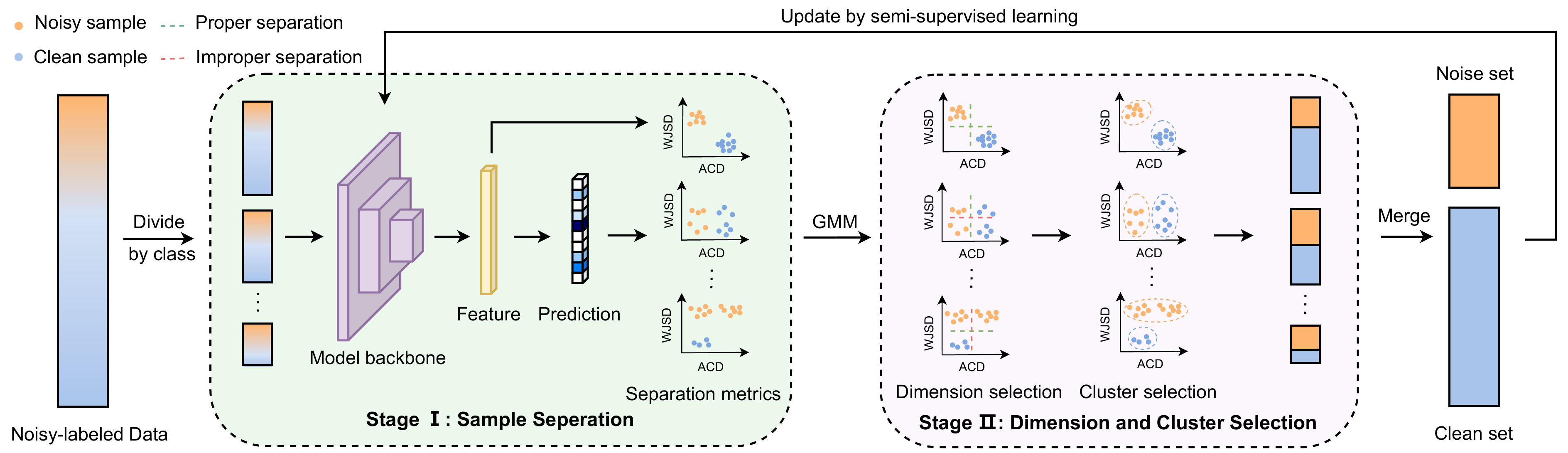}
\vspace{-10px}
\caption{The proposed framework for label-noise learning with intrinsically long-tailed data.
}
\label{framwork}
\vspace{-14px}
\end{figure*}

\section{Problem Definition}

Given a training set  $\DD = \{x_i, \hat{y}_i\}^{N}_{i=1} $, where $\hat{y}_i \in [M]$ is the observed label of the sample $x_i$.
$N$ is the number of training samples, and $M$ is the number of classes. 
In our problem, $\DD$ has the following properties:
\begin{itemize}
    \setlength{\itemsep}{2pt}
    \setlength{\parsep}{2pt}
    \setlength{\parskip}{2pt}
\item Noisy-labeled. There is a subset of samples $\hat{\DD} \in \DD$, where a sample $\{x,\hat{y}\} \in \hat{\DD}$ has an unknown ground-truth label $y$ different from its observed label $\hat{y}$.
\item Long-tailed. The ground-truth class distribution is long-tailed. Supposing $n_{c}$ is the number of the samples in class $c$, we have $n_1 > n_2 > \cdots > n_M$. 
\end{itemize}
We call this kind of training data \textit{intrinsically long-tailed} because the observed class distribution may not be long-tailed due to the existence of noisy labels. 
Therefore, our goal is to learn from noisy-labeled data with the intrinsically long-tailed class distribution.
The problem is challenging from two aspects.
On the one hand, directly applying long-tail learning methods \cite{DBLP:conf/cvpr/CuiJLSB19,DBLP:conf/nips/CaoWGAM19,DBLP:conf/cvpr/TanWLLOYY20} is infeasible because the observed class distribution may be inconsistent with the intrinsic class distribution.
On the other hand, label-noise learning methods perform poorly in the tail classes because they usually contain more noisy samples due to insufficient data, as shown in \cref{motivation}(a).
It results in high noise rates in the tail classes, which brings great challenges to separating clean samples and noisy samples for further training.
Under this circumstance, the existing sample selection methods \cite{DBLP:conf/iclr/LiSH20,Karim_2022_CVPR} often fail to select clean samples in the tail class. 
The main reason is that it is difficult to distinguish the tail class data from the noisy label data by existing metrics like cross-entropy loss, 
as shown in \cref{motivation}(b). 
\section{Proposed Method}
In this paper, we propose a two-stage bi-dimensional sample selection (TABASCO) to address the problem of label-noise learning with intrinsically long-tailed data.
TABASCO decouples the sample selection process into two stages: (1) sample separation, and (2) dimension and cluster selection.
First, given an initial model $\theta$ trained on the original training data $\DD$, we propose to calculate bi-dimension metrics for each sample in each observed class based on the outputs and features of the model.
Second, we determine the separation dimension with better separability for each class, and we adopt the corresponding selection strategy to select the cluster with more clean samples based on the selected metric.
Last, we adopt semi-supervised learning to update the model by regarding the selected clean cluster as labeled data.
The overall framework  is shown in \cref{framwork}.
\vspace{-4px}
\subsection{Bi-Dimensional Separation Metrics}
Due to the deficiency of a single separation metric to distinguish clean samples from noisy samples in the complex situation of noisy labels with intrinsic long-tail distribution, we propose to jointly use two metrics from different perspectives: weighted Jensen-Shannon divergence (WJSD) and adaptive centroid distance (ACD).
Both metrics are specifically designed for the joint problem and are complementary to cover the case when only one of them cannot separate clean samples from noisy samples well.
WJSD fully utilizes the information of prediction confidence, while ACD relies on the distance in the feature space. 
Thus, using clustering on samples according to the values of their bi-dimensional metrics has the flexibility to separate samples with different imbalance ratios, noise ratios, and noise types.

We first reduce the separation granularity to alleviate the interference from the head class to the tail class.
Specifically, the separation is performed within each observed class according to the proposed bi-dimensional metrics.
We first divide training set $\DD$ into subsets according to the observed labels $\DD_{c} = \{ (x,\hat{y}) \ | \ \hat{y} = c \} $. 
Given a sample $(x_i,\hat{y}_i)\in\DD_{c}$, the model predictions obtained from the model $\theta$ is denoted as $\mathbf{p}_i  = [p^{1}_i,p^{2}_i,...,p^{M}_i]$, where $p^{j}_i$ is the $j$'s dimension of vector $\mathbf{p}_i$.
The one-hot representation of the observed label $\hat{y}_i$ is denoted as $\hat{\mathbf{y}}_i$.

\vspace{4px}
\noindent
\textbf{Weighted Jensen-Shannon Divergence}
The Jensen-Shannon Divergence (JSD) is a commonly used metric to separate samples by assessing the variability of model predictions \cite{DBLP:conf/cvpr/YaoSZS00T21,Karim_2022_CVPR}. It is defined as:
\begin{align}\label{jsd}
JSD(x_i)
=\frac{1}{2} KL\Big(\mathbf{p}_i \Big\| \frac{\mathbf{p}_i +\hat{\mathbf{y}}_i}{2}\Big) 
+\frac{1}{2} KL\Big(\hat{\mathbf{y}}_i  \Big\|\frac{\mathbf{p}_i +\hat{\mathbf{y}}_i}{2}\Big),
\end{align}
where $KL(\cdot \| \cdot)$ is the Kullback-Leibler divergence.
When the intrinsic distribution is balanced, the JSD values of clean samples are generally lower than the noisy samples, so the separation can be easily modeled.
However, when the distribution is intrinsically long-tailed, the noisy and clean samples in a tail class tend to obtain similar prediction confidence on the observed class $c$, because their predictions are both towards the head classes.
In this case, using JSD as a separation metric fails to separate them.
The reason is given by \cref{theorem}, which shows the upper bound of the absolute value of the JSD difference between two samples. 
\vspace{-5px}
\begin{theorem}[Upper bound of JSD difference]
\label{theorem}
Suppose $x_i$ and $x_j$ are two samples in class $c$, $p_i^c$ and $q_i^c$ are the $c$'s dimension of their prediction confidence $\mathbf{p}_i=[p_i^1,p_i^2,...,p_i^M]$ and $\mathbf{p}_j=[p_j^1,p_j^2,...,p_j^M]$, respectively. 
The upper bound of the absolute value of the difference between their JSD values is given by:
\begin{align}\nonumber
|JSD(x_i)-JSD(x_j)|\le \frac{1}{2}\log\left(\frac{p_i^c+1}{p_i^c}\right)|p_i^c-p_j^c|.
\end{align}
\end{theorem}
Theorem 1 shows that the difference between values of JSD is only determined by the prediction confidence of the observed class, i.e., $p_i^c$ and $p_j^c$.
In addition, as $|p_i^c-p_j^c|$ gets smaller with a fixed value of $p_i^c$, $|JSD(x_i)-JSD(x_j)|$ is also smaller.
This indicates that when two samples are in the same class $c$ with close values of $p_i^c$,  their JSD values are also close, which makes it difficult to separate them.

Nevertheless, we can utilize the prediction confidence on other classes, i.e., $p_i^d$ for $d\ne c$, to distinguish two samples even if their values of $p_i^c$ are close.
Specifically, we propose WJSD by imposing an additional weight on JSD to further distinguish samples by inspecting their prediction confidence on other classes rather than the observed class. 
First, we take the maximum prediction confidence $\max(\mathbf{p}_i)$ into account because it may reflect the confidence of a noisy sample's ground-truth class $y_i$.
Then, we calculate the ratio of the maximum prediction confidence to the prediction confidence of the observed class $\max(\mathbf{p}_i)/p_i^c$.
In this manner, the additional weight is greater than one only if the class with the maximum prediction confidence is not the observed class.
The larger gap between $\max(\mathbf{p}_i)$ and $p_i^c$ makes the weight higher.
To avoid exceptionally large weights by the division during normalization over all samples in class $c$ for later clustering, we set the upper bound according to the averaged prediction confidence over all samples in class $c$. 
Finally, the additional weight can be calculated as follows:
\begin{align}\label{w}
W(x_i)=\min(\max(\mathbf{p}_i)/p_i^c,\max(\bar{\mathbf{p}}_c)/\bar{p}_c^c), 
\end{align} 
where $\bar{\mathbf{p}}_c=[\bar{p}_c^1,\bar{p}_c^2,...,\bar{p}_c^M]=\frac{1}{|\DD_c|}\sum_{i=1}^{|\DD_c|}\mathbf{p}_i$ is the average prediction confidence of class $c$.
Thus, WJSD can be calculated by:
\begin{align}\label{wjsd}
WJSD(x_i) = W(x_i)\times JSD(x_i).
\end{align} 

\textbf{Remarks}: Compared with the value of JSD that is only related to the prediction confidence on the observed class, adopting the maximum prediction confidence in WJSD can better separate clean samples from noisy samples. 

\vspace{4px}
\noindent
\textbf{Adaptive Centroid Distance. }
Although the sample separability of WJSD is greatly improved compared with JSD, the separation metric still relies on model prediction. 
When all the noisy samples labeled in a class are from another class, e.g., asymmetric noise, the model prediction for both noisy and clean samples will be highly similar because it is easy to learn a classifier that maps the features from two different classes into one.
It results in low discrimination between the model predictions of clean and noisy samples in this case.
Thus, solely using WJSD may not be enough to separate the clean and noisy samples when  model predictions (in terms of all classes, not only about $p^c$) are close.

Except for separating the samples in the output space, we propose another metric calculated in the feature space to eliminate the bias from the classifier because the learned features are more robust to noise labels.
Specifically, for a given sample, we can calculate the distance between its feature and the class feature centroid to evaluate how the feature of a sample deviates from its class centroid.
This approach is only practical when the quality of the centroid is high.
Directly calculating the centroid according to the observed label may not be accurate because it involves a certain number of features of the noisy samples.
Therefore, we define  purity in \cref{purity_d} to assess the quality of the centroid of class $c$.
\vspace{-5px}
\begin{definition}[Purity]
\label{purity_d}
Suppose a noisy sample set of class $c$ be $ \DD_{c} = \left \{ (x,\hat{y}) \ | \ \hat{y} = c \right \}$, where the intrinsic label corresponding to the sample is $y$. 
The purity $P_{ \DD_{c}}$ of set $ \DD_{c}$ is calculated by:
\begin{align}\label{purity}
P_{ \DD_{c}} = \max_{k=1,...,M}  \left \{  \frac{\sum_{i=1}^{N} I \left ( y_{i}=k \right )}{N} \right \},
 \end{align} 
where $I(\cdot )$ is an indicator function and $N$ is the  number of samples in the set $\DD_{c}$.
\end{definition}
\vspace{-5px}
Purity indicates the proportion of an intrinsic class that takes the majority in the observed class $c$.
The higher the purity of a class set used to calculate the centroid, the better the centroid helps to distinguish between noise and clean samples.
RoLT \cite{DBLP:journals/corr/abs-2108-11569} adopts a similar idea to use class feature centroid for noisy sample detection. 
However it directly calculates the feature centroid on the observed class, such that it inevitably suffers from the existence of noise features when the purity is low, especially in our problem.

In order to improve the purity of the class feature centroid for distance calculation between features, we propose ACD as the second separation metric to separate samples jointly with WJSD. 
The feature centroid for each class is adaptively updated by involving samples with high confidence from the observed class.
The class centroid $\mathbf{o}_{c}$ based on a high-confidence sample set $\DD^{H}_{c}$ is calculated by:
\begin{align}
\label{adc}
\mathbf{o}_{c} &= \frac{1}{\left | \DD_{c}^H \right | }\sum_{i=1}^{|\DD_{c}^H|}\mathbf{f}_i,\\
\label{hlc}
\DD^{H}_{c} &= \{ x_i  | x_i  \in \DD_{c},  \ p_i^{t_c}>H_{c}\},
\end{align}
where $\mathbf{f}_i$ is the feature of $x_i$, and $t_c=\argmax_c\{\bar{p}_c^j\}$ is the class index with the largest average prediction confidence of class $c$.
Thus, we can use the prediction confidence of class $t_c$ for sample $x_i$, e.g., $p_i^{t_c}$, as the selection criteria compare with the threshold $H_c$.
$\DD^{H}_{c}$ is constructed by the samples in $D_c$ whose corresponding prediction confidence of class $t_c$ is higher than $H_c$.
The high-confidence threshold $H_{c}$ is defined as:
\begin{align}
H_{c} &= \frac{1}{D_{c}} \sum_{i = 1}^{\left | D_{c} \right | } w_{i}\times p^{t_c}_{i},\\
w_{i}  &= \max\left ( 1, p^{t_c}_{i}/\bar{p}^{t_c}_c \right ). 
\end{align}
$H_{c}$ is calculated by the weighted sum of the prediction confidence of class $t_c$ for all samples $x_i$.
The weight $w_i$ increases when a sample's prediction confidence of class $t_c$ is higher than its class average.
Thus, the threshold is high enough to ensure the purity of the selected samples.
High confidence samples are more representative and more likely to be in the same class.
It is worth noting that we choose more representative samples rather than clean samples because the clean samples in the tail classes are easily overwhelmed by the noise samples in our problem. 
In this case, it is difficult to select the clean samples directly because the noise samples may be more representative in the tail class. 
As for asymmetric noise, the choice of clean or noisy samples to obtain centroid for sample separation is equivalent.

In each round, the centroids are adaptively adjusted because the feature extractor in the model is updated, as well as the set $\DD_c^H$.
Therefore, the proposed metric ACD can be calculated by:
\begin{align}\label{acd}
ACD(x_i)=\cos(\mathbf{f}_i,\mathbf{o}_{c}).
\end{align}

\textbf{Remarks}: The proposed two metrics, WJSD and ACD, are complementary for sample separation in the tail classes.
When the ground-truth classes of the noisy samples are diverse, e.g. symmetric noise, WJSD plays the dominant role because the predictions can hardly be unified due to the class diversity of noisy samples.
They are more likely to be predicted towards their ground-truth class, which can be captured by $\max(\mathbf{p}_i)$.
In this case, ACD may not be effective in separating clean samples from noisy samples because the messy noisy samples may affect the purity of the calculated centroid.
On the other hand, when the ground-truth classes of the noisy samples belong to only one class, e.g., asymmetric noise, WJSD tends to produce similar predictions for noisy and clean samples, while the adaptive centroid in ACD can help distinguish them in the feature space.

\subsection{Bi-Dimensional Sample Selection}\label{cluster_selection}
Once the values of the bi-dimensional metrics for all samples in class $c$ are calculated, each sample can be represented by a point in a 2D space.
For each dimension, Gaussian mixture model (GMM) can be adopted to separate the samples in class $c$ into two clusters $\mathcal{G}^{1}_{c}, \mathcal{G}^{2}_{c}$.

\begin{figure}[!b]
\vspace{-8px}
\includegraphics[width=\columnwidth]{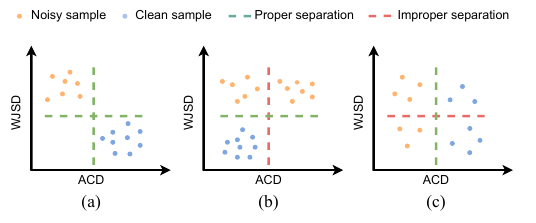}
\vspace{-20px}
\caption{Examples of three cases of the optimal separation dimension correspond to sample distribution in bi-dimensional metrics.}
\label{dimension_selection}
\vspace{-5px}
\end{figure}

\vspace{4px}
\noindent
\textbf{Dimension Selection.}
Because each metric shows different separability for different noise types, we first propose a dimension selection strategy to select a proper dimension to separate clean and noisy samples. 
We consider three cases to select the proper dimension: (a)  both of them are acceptable; (b) the separability of WJSD is better; (c) the separability of ACD is better.
In short, we only need to figure out if WJSD is suitable for sample separation.
If it is certain that WJSD does not show good separability, ACD will be adopted.
\cref{dimension_selection} shows three cases for dimension selection.
\begin{algorithm}[!t]\label{algorithm}
	\caption{The Dimension Selection Strategy}
	\label{alg:dimension_selection}
	\LinesNumbered 
	\KwIn{Noise sample set  $\DD_c$ in class $c$, threshold $\eta$}
	\KwOut{Cluster $\mathcal{G}^{1}_{c}$, $\mathcal{G}^{2}_{c}$}
	Obtain $\mathcal{G}^{1}_{wjsd}$, $\mathcal{G}^{2}_{wjsd}$ and separation threshold $d$ by applying GMM with values of WJSD  to all samples in $\DD_c$ \\
	Obtain $\mathcal{G}^{1}_{acd}$, $\mathcal{G}^{2}_{acd}$ by applying GMM with values of ACD  to all samples in $\DD_c$ \\
    Calculate the mean $\mu_1,\mu_2$ and variance $\sigma_1,\sigma_2$ of WJSD  for $\mathcal{G}^{1}_{acd}, \mathcal{G}^{2}_{acd}$ \\
    \uIf{$\mu_{1} < d < \mu_{2}  $  and $ \sigma_{2} / \sigma_{1} < \eta $ }{
     return $\mathcal{G}^{1}_{wjsd}, \mathcal{G}^{2}_{wjsd}$  
    }
    \uElseIf{$\mu_{1} > d $  and  $ \mu_{2} >d$  }{
    return $\mathcal{G}^{1}_{wjsd}, \mathcal{G}^{2}_{wjsd}$ 
    }
    \Else{
    return $\mathcal{G}^{1}_{acd}, \mathcal{G}^{2}_{acd}$ 
    }
\end{algorithm}
\setlength{\textfloatsep}{15pt}

Therefore, we use the statistics of WJSD as the criterion of dimension selection. 
We measure the means $(\mu_1, \mu_2)$ and standard deviations $(\sigma_1, \sigma_2)$ of WJSD for two clusters separated by GMM in terms of ACD. 
We have two cases that can determine when WJSD is better.
We mainly compare $\mu_1$ and $\mu_2$ with the threshold $d$ of GMM in WJSD.
First, as shown in \cref{dimension_selection} (b), if one cluster $\mathcal{G}^{1}_{acd}$ of ACD  has a $\mu_1$ less than $d$ and a $\sigma_1$ much larger than $\sigma_2$ in another cluster with  a $\mu_2$ larger than $d$, it means that  cluster $\mathcal{G}^{1}_{acd}$ has more clean samples but also many noise samples. So in this case, ACD is improper and WJSD is proper.
Second, if both $\mu_1, \mu_2$  are greater than $d$, it means that it is a representation of the previous case in the tail class because there are fewer clean samples in the cluster.
So, in this case,  WJSD is proper.
The strategy is summarized in Algorithm 1.

\vspace{4px}
\noindent
\textbf{Cluster Selection.}
Once we have determined the separation dimension, we can obtain two sample clusters $\mathcal{G}^{1}_{c}$ and $\mathcal{G}^{2}_{c}$.
However, it is not clear which cluster contains more clean samples because the noisy samples may overwhelm the clean samples in some cases.
Therefore, we need to further select a cluster between $\mathcal{G}^{1}_{c}$ and $\mathcal{G}^{2}_{c}$ as the clean one.
For the cluster separated based on WJSD, we simply choose the one with a smaller average WJSD value as the cluster with more clean samples, denoted as $\DD^{clean}_{c}$ \cite{DBLP:conf/iclr/LiSH20,Karim_2022_CVPR}.
For the cluster separated based on ACD, we cannot directly select the cluster closest to the centroid as the cluster with more clean samples since we take more representative samples, which may compose of noisy samples, rather than clean samples as the centroid.
Therefore, we need to determine the selection criteria based on whether the current class centroid is obtained by clean samples.
Especially, the criterion is based on the similarity between centroids of the current class and other classes.
Assume that the sample in $\mathcal{G}^{1}_{c}$ is near the centroid and the sample in $\mathcal{G}^{2}_{c}$ is the opposite.
We determined the choice of a more clean cluster $\DD^{clean}_{c}$  based on the following criteria:
\begin{align}
\label{cluster_selection_}
\DD^{clean}_{c}&=\begin{cases} \mathcal{G}^2_{c} , \  \text{if} \ |\cos\left ( o_c,o_k \right ) - 1| < \varepsilon    \  \text{and} \  \left | \DD^H_c \right | < \left | \DD^H_k\right |   
\\  \mathcal{G}^1_{c}, \  \text{otherwise}   
\end{cases} 
\end{align}
It means that if the centroid of another class is similar to the centroid of the current class, whichever class of high-confidence sample set has fewer samples, then the high-confidence samples in this class are the noise samples. 
It is because the asymmetric noise ratio cannot exceed  50\% in the past assumption \cite{DBLP:conf/iclr/LiSH20,Karim_2022_CVPR}.
Therefore, if the centroid is obtained from noise samples, we choose the cluster far away from the centroid as the cluster with more clean samples; otherwise, the cluster close to the centroid is selected.
\vspace{-16px}
\subsection{Overall Training Process}
After sample selection is conducted for each class, we adopt semi-supervised learning \cite{DBLP:conf/nips/BerthelotCGPOR19,DBLP:conf/nips/SohnBCZZRCKL20} to train with all clean samples $\DD^{clean}$ as labeled data and all noisy samples $\DD^{noisy}$ as unlabeled data.
The model is thus updated for the next round of training.
One may also use long-tailed semi-supervised learning methods instead of normal semi-supervised learning in this stage.
\vspace{-5pt}
\section{Experiments}
\subsection{Experimental Setup}
\noindent\textbf{Datasets.}
In order to comprehensively evaluate the effectiveness of  method, we perform experiments respectively in the scenarios with  synthetic noise and realistic noise in intrinsically long-tailed distribution.
We construct synthetic scenarios based on CIFAR-10 and CIFAR-100 \cite{Krizhevsky2009LearningML}.
We construct two benchmarks to imitate realistic scenarios, which is built on real-world noise datasets including Red Mini-ImageNet \cite{DBLP:conf/icml/JiangHLY20}, CIFAR-10N and CIFAR-100N \cite{wei2022learning}.
For all datasets, we adopt the standard paradigm of constructing a long-tailed distribution first and injecting noise later.
We imitate realistic long-tailed distribution using the same setting in previous long-tailed learning works \cite{DBLP:conf/nips/CaoWGAM19,Park_2021_ICCV} that long-tailed imbalance follows an exponential decay in sample sizes across different classes.
The imbalance factor is denoted by the ratio between the size of the largest class and that of the smallest class.

The details for dataset construction are as follows.
(1) As for CIFAR-10/100 \cite{Krizhevsky2009LearningML}, we adopt symmetric and asymmetric noise to inject synthetic noise, which is commonly used in the area of label-noise learning \cite{DBLP:conf/nips/ShuXY0ZXM19}.
The noise ratio is denoted by the ratio between the number of noisy samples and the total number of samples.
It should be noted that different from the previous methods \cite{jiang2022delving,DBLP:journals/corr/abs-2108-11569} to deal with the joint problem of noisy labeled and long-tailed data, the noise transition matrix is randomly generated only related to the noise ratio.
(2) As for CIFAR-10N/100N (10N/100N) \cite{wei2022learning}, we replace sample labels with human-annotated noisy labels after constructing a long-tailed distribution based on real labels.
(3) As for Red mini-ImageNet (Red) \cite{DBLP:conf/icml/JiangHLY20}, we inject web label noise samples after constructing a long-tailed distribution based on the original mini-ImageNet.
Therefore, the observed and intrinsic distribution are likely inconsistent, especially in long-tail distribution.

\vspace{2px}
\noindent\textbf{Compared Methods.}
We compare our method with the following three types of approaches:
(1) Long-tail learning methods (LT). They are LA \cite{DBLP:conf/iclr/MenonJRJVK21}, LDAM \cite{DBLP:conf/nips/CaoWGAM19} and IB \cite{Park_2021_ICCV}.
(2) Label-noise learning methods (NL). They are  DivideMix \cite{DBLP:conf/iclr/LiSH20} and UNICON \cite{Karim_2022_CVPR};
(3) Methods aiming at dealing with noisy label and long-tail distribution (NL-LT). They are MW-Net \cite{DBLP:conf/nips/ShuXY0ZXM19}, RoLT \cite{DBLP:journals/corr/abs-2108-11569}, HAR \cite{DBLP:conf/iclr/CaoCLAGM21} and ULC \cite{DBLP:conf/aaai/HuangBZBW22}.

\vspace{2px}
\noindent\textbf{Implementation Details.}
We use PreAct ResNet18 \cite{DBLP:conf/cvpr/HeZRS16} as backbone for CIFAR datasets and  ResNet18 as backbone for Red mini-ImageNet dataset.
Both of backbones adopt SGD with an initial learning rate of 0.02, a momentum of 0.9, and a weight decay of $5\times 10^{-4}$ and a batch size of 64.
For fair comparison,  we train all methods for 100 epochs.
\begin{table}[!t]\scriptsize
        \centering
        \resizebox{\columnwidth}{!}{
        \begin{tabular}{@{}l|llllcllll@{}}
                \toprule
                \multicolumn{2}{l}{Dataset}       & \multicolumn{2}{c}{CIFAR-10}      & \multicolumn{2}{c}{CIFAR-100}       \\ 
                \midrule
                \multicolumn{2}{l}{Imbalance Factor}       & \multicolumn{4}{c}{0.1}     \\
                \midrule
                \multicolumn{2}{l}{Noise Ratio (\textbf{Sym.})}  & 
                \multicolumn{1}{c}{0.4} & \multicolumn{1}{c}{0.6} & \multicolumn{1}{c}{0.4} & \multicolumn{1}{c}{0.6}\\ \midrule
                Baseline & CE & {71.67} & {61.16}   & {34.53} & {23.63}    \\ \midrule
                \multirow{3}{*}{LT}& LA   & {70.56} & {54.92}   & {29.07} & {23.21}    \\
                 & LDAM  & {70.53} & {61.97}   & {31.30} & {23.13}    \\
                 & IB  & {73.24} & {62.62}   & {32.40} & {25.84}    \\  \midrule
                \multirow{2}{*}{NL}& DivideMix     & {82.67} & {80.17}    & {54.71} & {44.98}  \\
                & UNICON      & {84.25} & {82.29}    & {52.34} & {45.87}  \\\midrule
                \multirow{4}{*}{NL-LT}& MW-Net  & {70.90} & {59.85}   & {32.03} & {21.71} \\
                &RoLT      & {81.62} & {76.58}   & {42.95} & {32.59} \\
                &HAR      & {77.44} & {63.75}    & {38.17} & {26.09} \\ 
                &ULC      & {84.46} & {83.25}    & {54.91} & {44.66} \\  \midrule

                \multirow{1}{*}{Our}&TABASCO     & {\textbf{85.53}} & {\textbf{84.83}}  & {\textbf{56.52}} & {\textbf{45.98}}  \\
               \bottomrule
        \end{tabular}
        }
        \caption{Performance comparison under symmetric noise. The best results are shown in bold.}
        \label{cifar_unif}
        \vspace{-10px}
\end{table}

\subsection{Comparative Results}
\noindent\textbf{CIFAR-10/100.}
\cref{cifar_unif} and \ref{cifar_flip} report the accuracy of different methods for intrinsically long-tailed CIFAR-10/100 with symmetric and asymmetric noise, respectively. 
It can be observed three phenomena as follows:
(1) The performance of existing long-tail methods is lower than baseline in most situations.
It is because such methods do not have the ability to distinguish the noise samples, and the wrong attention to the noise samples leads to the further decline of the model generalization ability.
(2) The advantages of existing methods aiming at both long-tailed distribution and  label noise are not obvious compared with single methods.
It is because existing methods do not take into account the distribution inconsistency caused by noise labels.
(3) Our method significantly improved over other methods in all cases.
Moreover, as the proportion of noise increases, the accuracy of our method decreases less compared to other methods.
It validates that our method can effectively identify the noise in long-tail distribution.

\begin{table}[!t]\scriptsize
        \centering
        \resizebox{\columnwidth}{!}{
        \begin{tabular}{@{}l|llllcllll@{}}
                \toprule
                \multicolumn{2}{l}{Dataset}         & \multicolumn{2}{c}{CIFAR-10}      & \multicolumn{2}{c}{CIFAR-100}       \\ 
                \midrule
                \multicolumn{2}{l}{Imbalance Factor}       & \multicolumn{4}{c}{0.1}     \\
                \midrule
                \multicolumn{2}{l}{Noise Ratio (\textbf{Asym.})}  & 
                \multicolumn{1}{c}{0.2} & \multicolumn{1}{c}{0.4} & \multicolumn{1}{c}{0.2} & \multicolumn{1}{c}{0.4}\\ \midrule
                Baseline & CE & {79.90} & {62.88} & {44.45} & {32.05}\\ \midrule
                 \multirow{3}{*}{LT}& LA   & {71.49} & {59.88}   & {39.34} & {28.49}    \\
                 & LDAM  & {74.58} & {62.29}   & {40.06} & {33.26}    \\
                 & IB  & {73.49} & {58.36}   & {45.02} & {35.25}    \\  \midrule
                 \multirow{2}{*}{NL}&  DivideMix    & {80.92} & {69.35} & {58.09} & {41.99}\\
                 & UNICON      & {72.81} & {69.04} & {55.99} & {44.70}\\\midrule
                \multirow{4}{*}{NL-LT}&  MW-Net  & {79.34} & {65.49} & {42.52} & {30.42}\\
                 &RoLT      & {73.30} & {58.29} & {48.19} & {39.32}\\
                &  HAR       & {82.85} & {69.19} & {48.50} & {33.20}\\
                &ULC      & {74.07} & {73.19}    & {54.45} & {43.20} \\  \midrule
                 \multirow{1}{*}{Our}& TABASCO     & {\textbf{82.10}} & {\textbf{80.57}} & {\textbf{59.39}} & {\textbf{50.51}}\\ 
                 \bottomrule
        \end{tabular}
        }
        \caption{Performance comparison under asymmetric noise. The best results are shown in bold.}
        \label{cifar_flip}
        \vspace{-10px}
\end{table}
\vspace{3px}
\noindent\textbf{Red mini-ImageNet and CIFAR-10N/100N.}
\cref{realist_datasets}  reports the accuracy of different methods on real-world noise datasets with intrinsically long-tailed distribution.
According to the results of label-noise learning methods on CIFAR-10N/100N, it can be observed that real-world noise is more difficult to separate than synthetic noise to some extent.
It further increases the difficulty of dealing with such problem. 
Nevertheless, our method still performs well, which again confirms its effectiveness.
\vspace{-2px}
\subsection{Ablation Study and Discussions}\label{ablation}
In the following experiments, we analyze each component of the proposed TABASCO on CIFAR-10/100 to verify its effectiveness.
We use an imbalance factor of 0.1 and  a noise ratio of 0.4 to construct the synthetic noise dataset.

\vspace{2px}
\noindent
\textbf{Effectiveness of WJSD.}
To validate the effectiveness of the proposed WJSD, we compare the accuracy of models trained by different separation metrics.
Specifically, we train the model by using JSD and WJSD for sample separation with a symmetric noise dataset, respectively.
As for each separation metric, we simply use the small value strategy \cite{DBLP:conf/iclr/LiSH20,Karim_2022_CVPR} for sample selection.
As \cref{wjsd_effectivenes} shows, the model trained by WJSD achieves better results on both symmetric noise datasets compared with JSD owing to the stronger separability of WJSD.
\begin{table}[!t]\small
\vspace{2px}
        \centering
        \begin{tabular}{@{}lllllcllll@{}}
                \toprule
                Dataset       & {Separation metric}          & {Accuracy}   \\ \midrule
                \multirow{2}{*}{CIFAR-10}& \makecell[c]{JSD}   & \makecell[c]{83.97}\\ 
                                          &  \makecell[c]{WJSD}  & \makecell[c]{85.57} \\ 
                                           \midrule
                \multirow{2}{*}{CIFAR-100}  & \makecell[c]{JSD}   & \makecell[c]{55.80}\\
                                            &   \makecell[c]{WJSD}  & \makecell[c]{56.67} \\
                \bottomrule 
        \end{tabular}
        \caption{Performance comparison between JSD and WJSD.}
        \label{wjsd_effectivenes}
        \vspace{-10px}
\end{table}

\begin{figure}[!t]

\begin{subfigure}[b]{0.47\columnwidth}
    \centering
    \includegraphics[width=\textwidth]{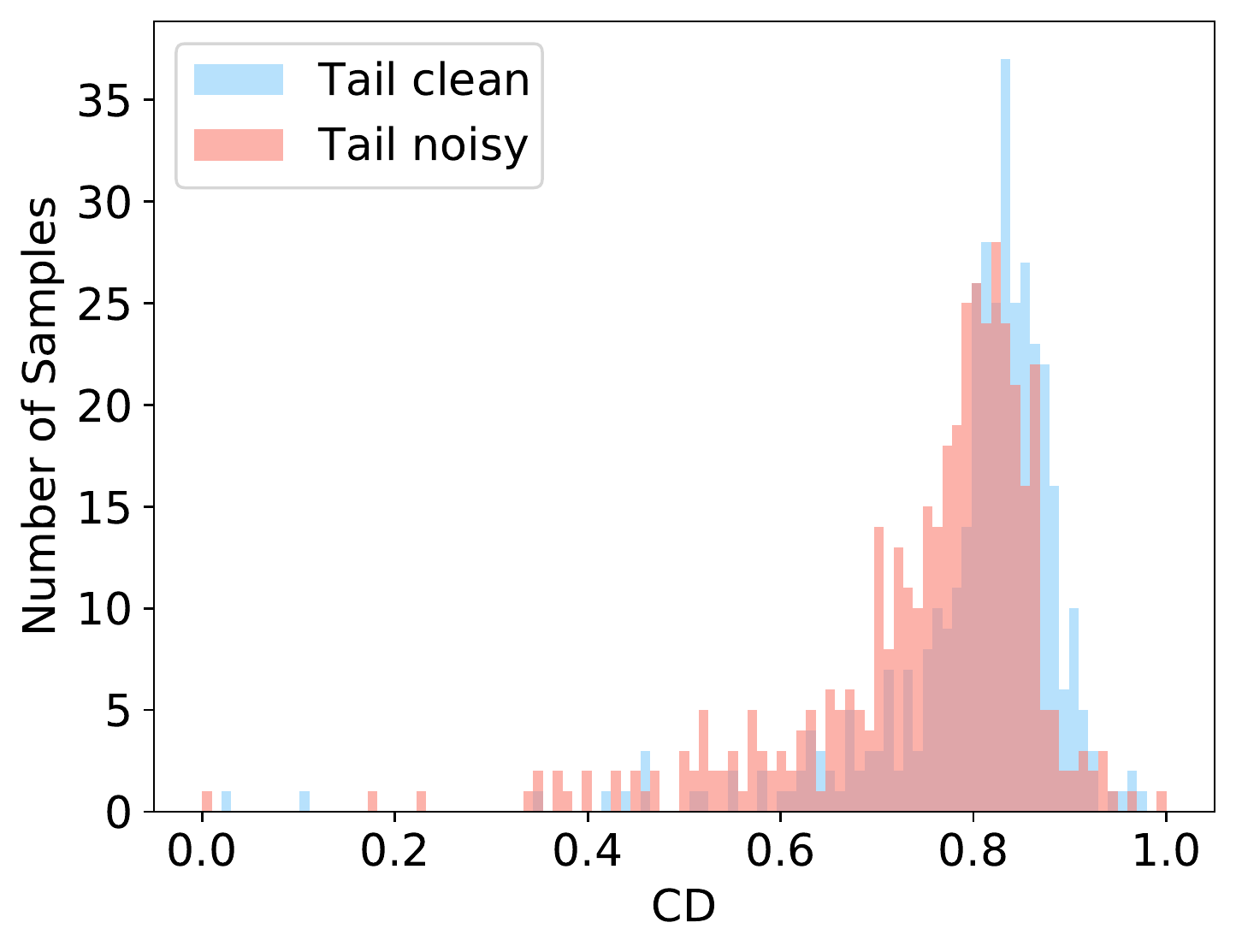}
    \vspace{-15px}
\end{subfigure}\hspace{1px}
\begin{subfigure}[b]{0.47\columnwidth}
    \centering
    \includegraphics[width=\textwidth]{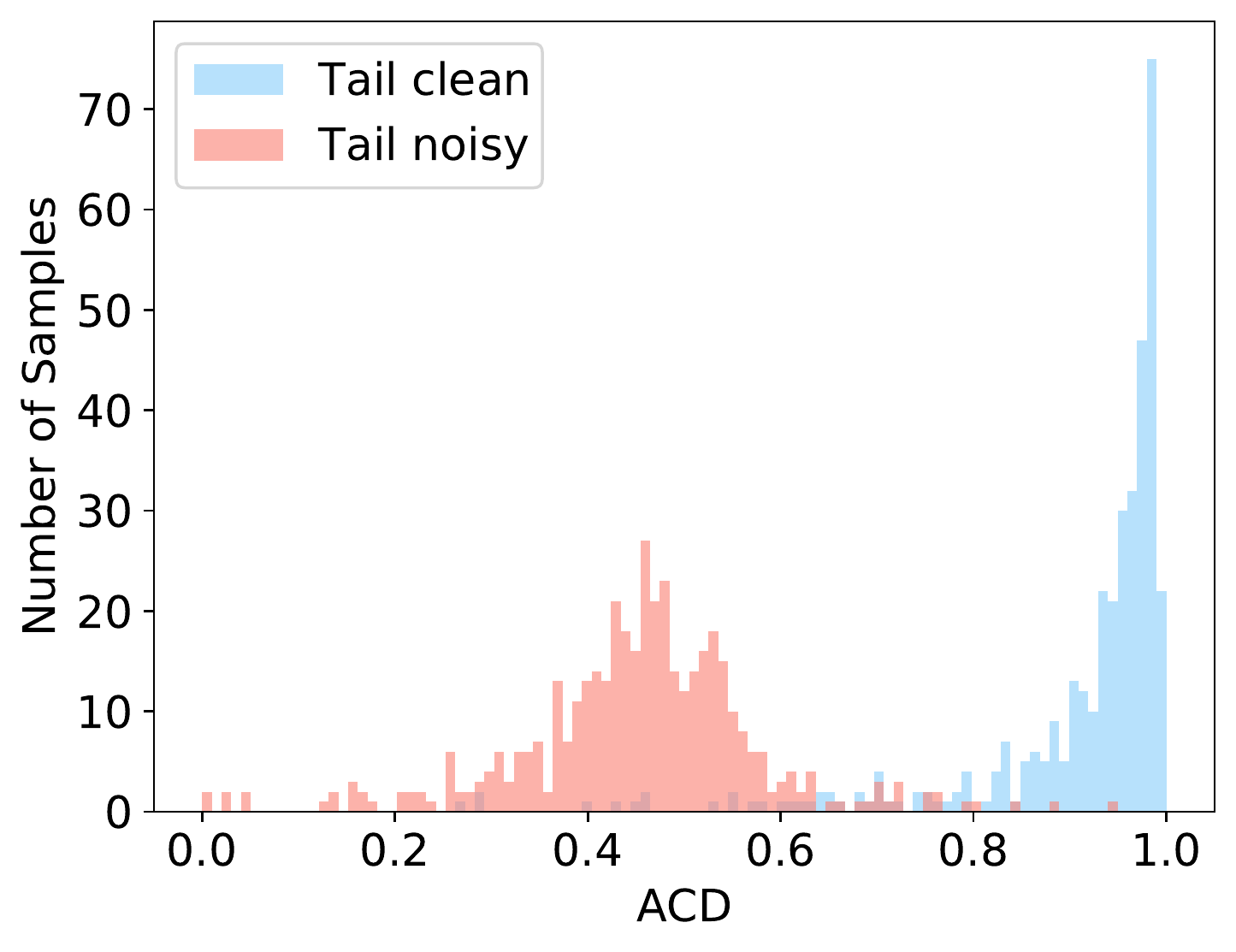}
    \vspace{-15px}
\end{subfigure}
\vspace{-5px}
\caption{Comparison of the sample distributions between CD and ACD in the tail class.}
\label{acd_fig}
\vspace{-12px}
\end{figure}

\vspace{2px}
\noindent
\textbf{Effectiveness of ACD.}
To validate the advantage of the proposed ACD, we compare it with the method which calculates the feature centroid on the observed class (hereinafter referred to as centroid distance or CD) to show their separability in CIFAR-10.
Specifically, based on the same backbone, we calculate the distance from the sample to the centroid using ACD and CD in CIFAR-10 with asymmetric noise, respectively.
\cref{acd_fig} shows the distributions of samples in the tail class by CD and ACD. 
The purity of the tail class is 0.53.
It can be observed that the CD values of noise and clean samples are highly overlapped, which leads to failure sample separation. 
It is because  the purity of the tail class may be quite low due to the negative effects of both label-noise and long-tailed distribution.
The centroid obtained by ACD has a higher purity, so there is a significant difference in ACD between them.
As shown in \cref{acd_fig}(b), most clean samples are concentrated in the interval of [0.8,1], which is easily clustered by GMM.
\begin{table}[!t]\scriptsize
        \centering
        \resizebox{\columnwidth}{!}{
        \begin{tabular}{@{}l|llllcllll@{}}
                \toprule
                \multicolumn{2}{l}{Dataset}  & \multicolumn{2}{c}{Red}   & \multicolumn{1}{c}{10N} & \multicolumn{1}{c}{100N}      \\ 
                \midrule
                \multicolumn{2}{l}{Imbalance Factor}    & \multicolumn{2}{c}{$\approx$ 0.1}   & \multicolumn{2}{c}{0.1}     \\
                \midrule
                \multicolumn{2}{l}{Noise Ratio}  & 
                \multicolumn{1}{c}{0.2} & \multicolumn{1}{c}{0.4} & \multicolumn{2}{c}{$\approx$ 0.4} \\ \midrule
                Baseline & CE & {40.42} & {31.46} & {60.44} & {38.10}\\ \midrule
                 \multirow{3}{*}{LT}& LA   & {26.82} & {25.88}   & {65.74} & {36.50}    \\
                 & LDAM  & {26.64} & {23.46}   & {62.50} & {38.48}    \\
                 & IB  & {23.80} & {22.08}   & {65.91} & {42.48}    \\  \midrule
                 \multirow{2}{*}{NL}&  DivideMix    & {48.76} & {48.96} & {67.85} & {44.25}\\
                 & UNICON      & {40.18} & {41.64} & {69.54} & {51.93}\\\midrule
                \multirow{4}{*}{NL-LT}&  MW-Net  & {42.66} & {40.26} & {69.73} & {44.20}\\
                 &RoLT      & {22.56} & {24.22} & {75.24} & {46.61}\\
                &  HAR       & {46.61} & {38.71} & {74.97} & {44.54}\\
                &ULC      & {48.12} & {47.06}    & {75.71} & {51.72} \\  \midrule
                 \multirow{1}{*}{Our}& TABASCO   & {\textbf{50.20}} & {\textbf{49.68}} & {\textbf{80.61}} & {\textbf{53.83}}\\ 
                 \bottomrule
        \end{tabular}
        }
        \caption{Performance comparison with real-world noise and long-tail distribution.
The best results are shown in bold.}
        \label{realist_datasets}
        \vspace{-8px}
\end{table}

\begin{figure}[!t]
\hspace{3px}
\begin{subfigure}[b]{0.45\columnwidth}
    \includegraphics[width=\textwidth]{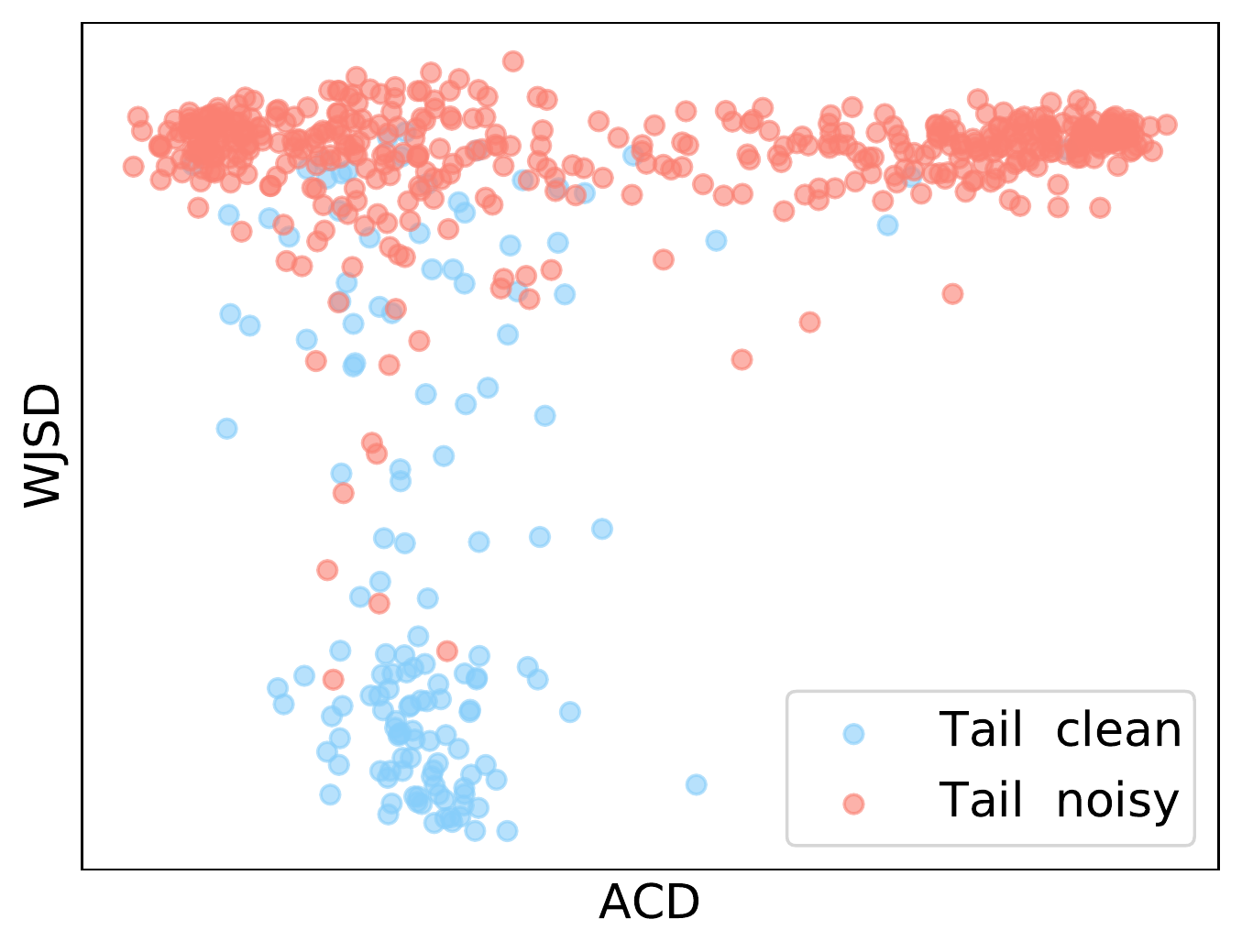}
    \vspace{-15px}
    \subcaption{Symmetric noise}
\end{subfigure}\hspace{4px}
\begin{subfigure}[b]{0.45\columnwidth}
    \includegraphics[width=\textwidth]{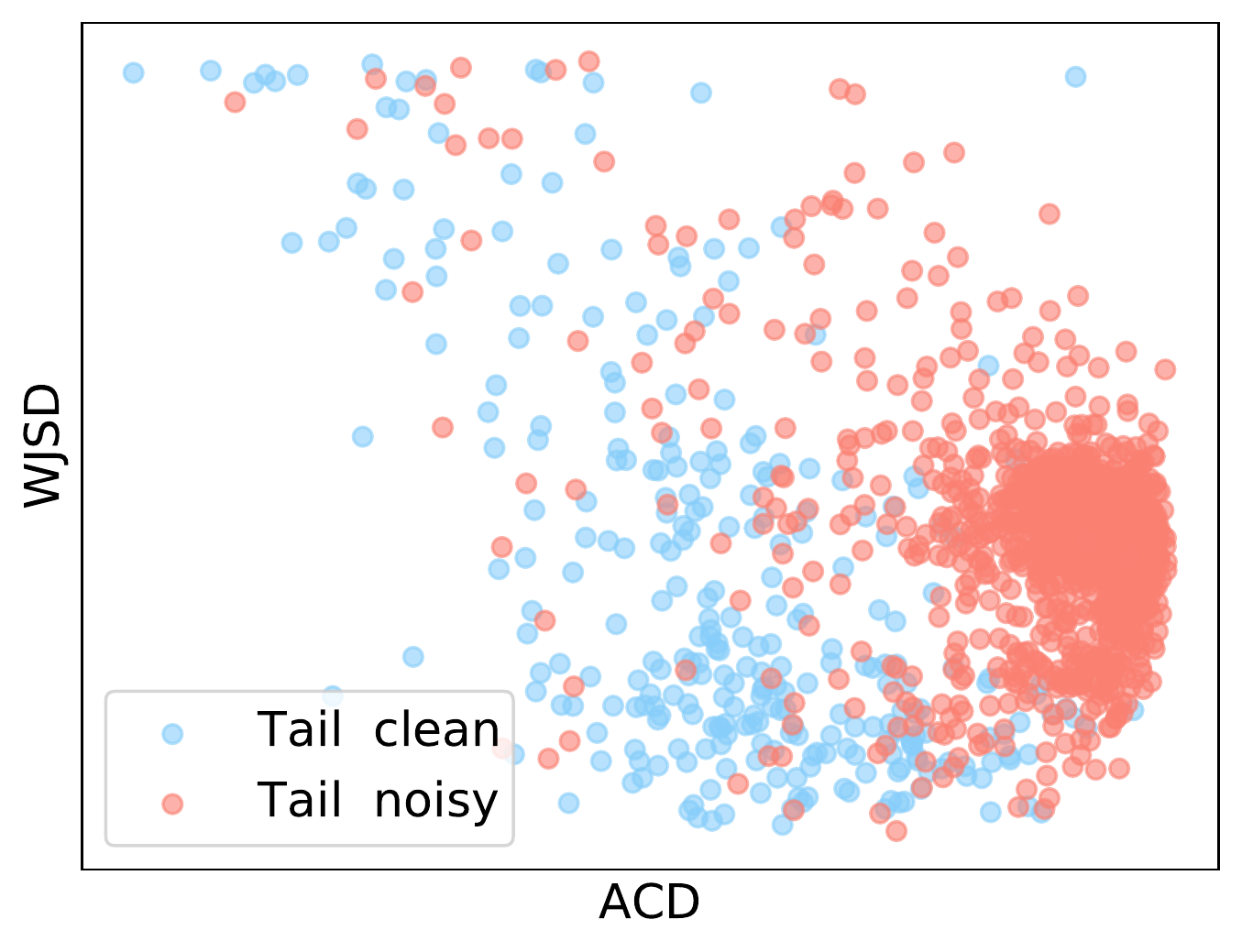}
    \vspace{-15px}
    \subcaption{Asymmetric noise}
\end{subfigure}
\vspace{-5px}
\caption{Scatter plot of the values of the proposed bi-dimensional metrics with (a) symmetric noise and (b) asymmetric noise.}
\label{bi-dimensional_metric}
\vspace{-15px}
\end{figure}

\vspace{2px}
\noindent
\textbf{Effectiveness of Bi-dimensional Sample Separation.}
To validate how the proposed bi-dimensional metrics complement each other, we plot the values of bi-dimensional metrics of both clean and noisy samples under different noise types in \cref{bi-dimensional_metric}.
For the case of symmetric noise shown in \cref{bi-dimensional_metric}(a), it can be observed that ACD cannot effectively distinguish clean samples from noisy samples in the tail class.
In this case, WJSD shows its advantage because the values of WJSD for most 
noisy samples are clustered in the top region, and the clean samples are scattered throughout the rest region.
For the case of asymmetric noise shown in \cref{bi-dimensional_metric}(b), it can be observed that WJSD cannot distinguish clean samples from noisy samples well for the tail class, while ACD can distinguish them well although the clean and noisy samples are closer in the tail class.
The experimental observation is consistent with the previous discussion, which validates the complementarity of bi-dimensional metrics.

\noindent
\textbf{Effectiveness of Sample Selection.}
In order to validate the effectiveness of the proposed sample selection in dimension selection and cluster selection, we compare the optimal clean ratio of clusters obtained by WJSD, ACD and our selection method in CIFAR-10 with different noise types.
The clean ratio of a cluster is calculated by the proportion of clean samples in the cluster to the total number of cluster samples.
As shown in \cref{cluster_selection}, on the tail class, the clean ratio of the cluster obtained by our selection method is aligned with the higher one between WJSD and ACD for both noise cases.
It shows that we can choose the most appropriate separation dimension in different situations.
Besides, the clean ratio of the cluster we selected is consistent with that of the optimal cluster in the corresponding dimension, which is sufficient to show that the method we proposed can select clean samples.
\begin{figure}[!t]
\begin{subfigure}[b]{0.47\columnwidth}
    \includegraphics[width=\textwidth]{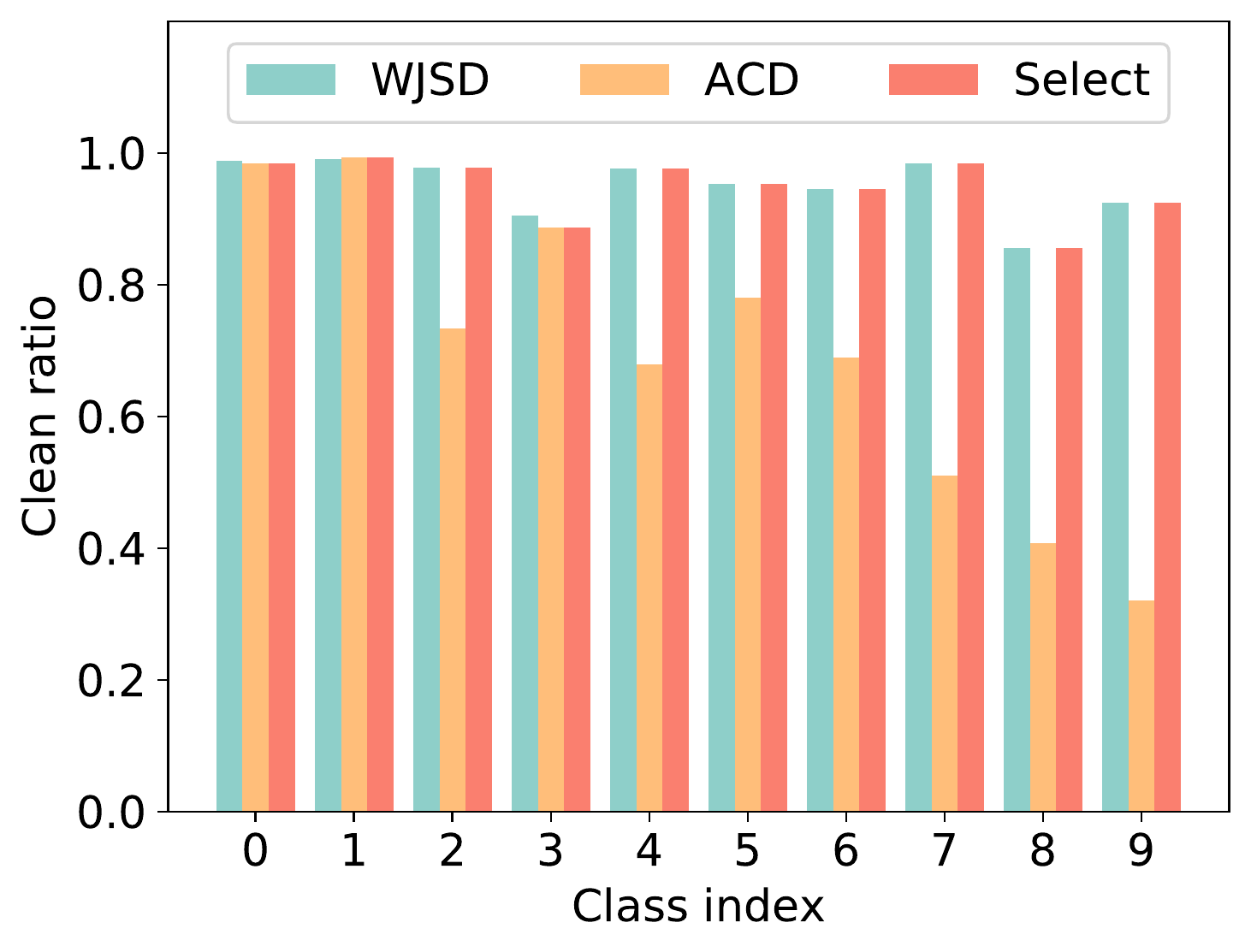}
    \subcaption{Symmetric noise}
\end{subfigure}\hspace{1px}
\begin{subfigure}[b]{0.47\columnwidth}
    \includegraphics[width=\textwidth]{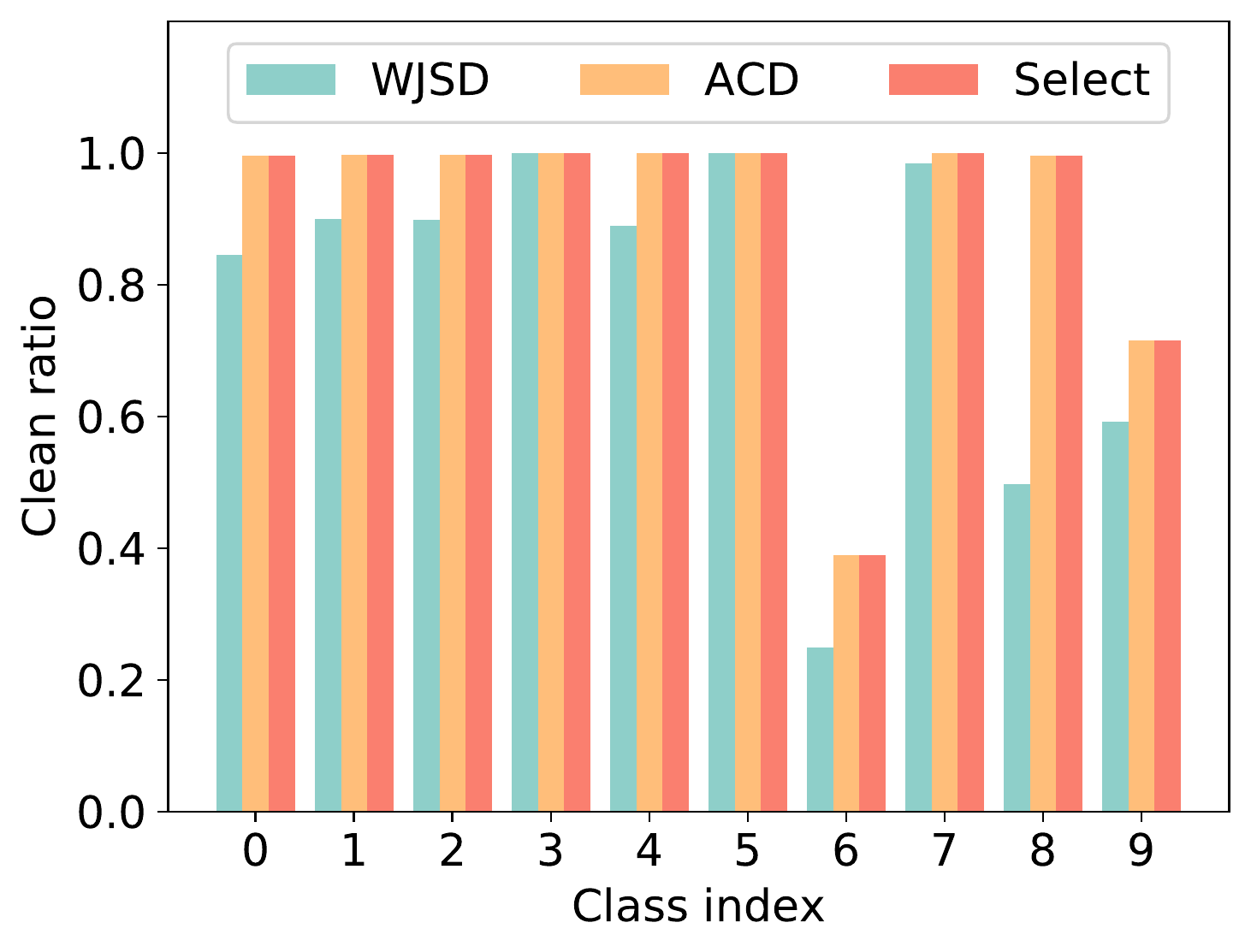}
    \subcaption{Asymmetric noise}
\end{subfigure}
\vspace{-5px}
\caption{The performance of the proposed sample selection with (a) symmetric noise and (b) asymmetric noise.}
\label{cluster_selection}
    \vspace{-10px}
\end{figure}
\vspace{-2pt}
\section{Conclusion}
\vspace{-3pt}
This paper studies a more general and realistic problem of label-noise learning with intrinsically long-tailed data.
The major challenge in this problem is that it is hard to distinguish clean samples from noisy samples on intrinsic tail classes.
Accordingly, we propose a learning framework TABASCO for this problem.
In TABASCO, two new metrics are explicitly proposed to address the problem of sample selection in tail classes.
Extensive experiments on noisy-labeled datasets with long-tail distribution demonstrate its effectiveness.

\noindent\textbf{Acknowledgements.}
We thank Dr. Shiquan Ruan for his discussion on the theory proof. 
This study was supported in part by the National Natural Science Foundation of China under Grants 62002302 and U21A20514; in part by the FuXiaQuan National Independent Innovation Demonstration Zone Collaborative Innovation Platform under Grant 3502ZCQXT2022008; in part by NSFC/RGC Joint Research Scheme under Grant N\_HKBU214/21; and in part by the General Research Fund of RGC under Grants 12201321 and 12202622.
{\small
\bibliographystyle{ieee_fullname}
\bibliography{egbib}

\begin{thebibliography}{10}\itemsep=-1pt

\bibitem{DBLP:conf/nips/BerthelotCGPOR19}
David Berthelot, Nicholas Carlini, Ian~J. Goodfellow, Nicolas Papernot, Avital
  Oliver, and Colin Raffel.
\newblock Mixmatch: {A} holistic approach to semi-supervised learning.
\newblock In {\em NeurIPS}, 2019.

\bibitem{DBLP:conf/iclr/CaoCLAGM21}
Kaidi Cao, Yining Chen, Junwei Lu, Nikos Ar{\'{e}}chiga, Adrien Gaidon, and
  Tengyu Ma.
\newblock Heteroskedastic and imbalanced deep learning with adaptive
  regularization.
\newblock In {\em ICLR}, 2021.

\bibitem{DBLP:conf/nips/CaoWGAM19}
Kaidi Cao, Colin Wei, Adrien Gaidon, Nikos Ar{\'{e}}chiga, and Tengyu Ma.
\newblock Learning imbalanced datasets with label-distribution-aware margin
  loss.
\newblock In {\em NeurIPS}, 2019.

\bibitem{DBLP:journals/jair/ChawlaBHK02}
Nitesh~V. Chawla, Kevin~W. Bowyer, Lawrence~O. Hall, and W.~Philip Kegelmeyer.
\newblock {SMOTE:} synthetic minority over-sampling technique.
\newblock {\em J. Artif. Intell. Res.}, 2002.

\bibitem{DBLP:conf/eccv/ChouCPWJ20}
Hsin{-}Ping Chou, Shih{-}Chieh Chang, Jia{-}Yu Pan, Wei Wei, and Da{-}Cheng
  Juan.
\newblock Remix: Rebalanced mixup.
\newblock In {\em ECCV}, 2020.

\bibitem{DBLP:conf/cvpr/CuiJLSB19}
Yin Cui, Menglin Jia, Tsung{-}Yi Lin, Yang Song, and Serge~J. Belongie.
\newblock Class-balanced loss based on effective number of samples.
\newblock In {\em CVPR}, 2019.

\bibitem{DBLP:conf/naacl/DevlinCLT19}
Jacob Devlin, Ming-Wei Chang, Kenton Lee, and Kristina Toutanova.
\newblock {BERT}: Pre-training of deep bidirectional transformers for language
  understanding.
\newblock In {\em NAACL}, 2019.

\bibitem{DBLP:journals/ci/EstabrooksJJ04}
Andrew Estabrooks, Taeho Jo, and Nathalie Japkowicz.
\newblock A multiple resampling method for learning from imbalanced data sets.
\newblock {\em Comput. Intell.}, 2004.

\bibitem{gupta2019lvis}
Agrim Gupta, Piotr Dollar, and Ross Girshick.
\newblock {LVIS}: A dataset for large vocabulary instance segmentation.
\newblock In {\em CVPR}, 2019.

\bibitem{DBLP:conf/icml/00030YYXTS20}
Bo Han, Gang Niu, Xingrui Yu, Quanming Yao, Miao Xu, Ivor~W. Tsang, and Masashi
  Sugiyama.
\newblock {SIGUA:} forgetting may make learning with noisy labels more robust.
\newblock In {\em ICML}, 2020.

\bibitem{DBLP:journals/corr/abs-2011-04406}
Bo Han, Quanming Yao, Tongliang Liu, Gang Niu, Ivor~W. Tsang, James~T. Kwok,
  and Masashi Sugiyama.
\newblock A survey of label-noise representation learning: Past, present and
  future.
\newblock {\em CoRR}, 2020.

\bibitem{DBLP:conf/nips/HanYYNXHTS18}
Bo Han, Quanming Yao, Xingrui Yu, Gang Niu, Miao Xu, Weihua Hu, Ivor~W. Tsang,
  and Masashi Sugiyama.
\newblock Co-teaching: Robust training of deep neural networks with extremely
  noisy labels.
\newblock In {\em NeurIPS}, 2018.

\bibitem{DBLP:conf/icic/HanWM05}
Hui Han, Wenyuan Wang, and Binghuan Mao.
\newblock Borderline-smote: {A} new over-sampling method in imbalanced data
  sets learning.
\newblock In {\em ICIC 2005}, 2005.

\bibitem{DBLP:conf/cvpr/HeZRS16}
Kaiming He, Xiangyu Zhang, Shaoqing Ren, and Jian Sun.
\newblock Deep residual learning for image recognition.
\newblock In {\em CVPR}, 2016.

\bibitem{DBLP:conf/iccv/HeWW21}
Yin{-}Yin He, Jianxin Wu, and Xiu{-}Shen Wei.
\newblock Distilling virtual examples for long-tailed recognition.
\newblock In {\em ICCV}, 2021.

\bibitem{DBLP:conf/nips/HendrycksMWG18}
Dan Hendrycks, Mantas Mazeika, Duncan Wilson, and Kevin Gimpel.
\newblock Using trusted data to train deep networks on labels corrupted by
  severe noise.
\newblock In {\em NeurIPS}, 2018.

\bibitem{DBLP:conf/cvpr/HornASCSSAPB18}
Grant~Van Horn, Oisin~Mac Aodha, Yang Song, Yin Cui, Chen Sun, Alexander
  Shepard, Hartwig Adam, Pietro Perona, and Serge~J. Belongie.
\newblock The inaturalist species classification and detection dataset.
\newblock In {\em CVPR}, 2018.

\bibitem{DBLP:journals/corr/abs-1709-01450}
Grant~Van Horn and Pietro Perona.
\newblock The devil is in the tails: Fine-grained classification in the wild.
\newblock {\em CoRR}, 2017.

\bibitem{DBLP:conf/aaai/HuangBZBW22}
Yingsong Huang, Bing Bai, Shengwei Zhao, Kun Bai, and Fei Wang.
\newblock Uncertainty-aware learning against label noise on imbalanced
  datasets.
\newblock In {\em AAAI}, 2022.

\bibitem{DBLP:conf/icml/JiangHLY20}
Lu Jiang, Di Huang, Mason Liu, and Weilong Yang.
\newblock Beyond synthetic noise: Deep learning on controlled noisy labels.
\newblock In {\em ICML}, 2020.

\bibitem{jiang2022delving}
Shenwang Jiang, Jianan Li, Ying Wang, Bo Huang, Zhang Zhang, and Tingfa Xu.
\newblock Delving into sample loss curve to embrace noisy and imbalanced data.
\newblock In {\em AAAI}, 2022.

\bibitem{DBLP:conf/iclr/KangXRYGFK20}
Bingyi Kang, Saining Xie, Marcus Rohrbach, Zhicheng Yan, Albert Gordo, Jiashi
  Feng, and Yannis Kalantidis.
\newblock Decoupling representation and classifier for long-tailed recognition.
\newblock In {\em ICLR}, 2020.

\bibitem{Karim_2022_CVPR}
Nazmul Karim, Mamshad~Nayeem Rizve, Nazanin Rahnavard, Ajmal Mian, and Mubarak
  Shah.
\newblock Unicon: Combating label noise through uniform selection and
  contrastive learning.
\newblock In {\em CVPR}, 2022.

\bibitem{DBLP:journals/corr/abs-2108-11096}
Shyamgopal Karthik, J{\'{e}}r{\^{o}}me Revaud, and Chidlovskii Boris.
\newblock Learning from long-tailed data with noisy labels.
\newblock {\em CoRR}, 2021.

\bibitem{DBLP:conf/nips/KimHPYHS20}
Jaehyung Kim, Youngbum Hur, Sejun Park, Eunho Yang, Sung~Ju Hwang, and Jinwoo
  Shin.
\newblock Distribution aligning refinery of pseudo-label for imbalanced
  semi-supervised learning.
\newblock In {\em NeurIPS}, 2020.

\bibitem{Krizhevsky2009LearningML}
Alex Krizhevsky.
\newblock Learning multiple layers of features from tiny images.
\newblock 2009.

\bibitem{DBLP:conf/nips/KrizhevskySH12}
Alex Krizhevsky, Ilya Sutskever, and Geoffrey~E. Hinton.
\newblock Imagenet classification with deep convolutional neural networks.
\newblock In {\em NeurIPS}, 2012.

\bibitem{DBLP:conf/nips/LeeSK21}
Hyuck Lee, Seungjae Shin, and Heeyoung Kim.
\newblock {ABC:} auxiliary balanced classifier for class-imbalanced
  semi-supervised learning.
\newblock In {\em NeurIPS}, 2021.

\bibitem{DBLP:conf/iclr/LiSH20}
Junnan Li, Richard Socher, and Steven C.~H. Hoi.
\newblock Dividemix: Learning with noisy labels as semi-supervised learning.
\newblock In {\em ICLR}, 2020.

\bibitem{Li_2022_CVPR}
Shikun Li, Xiaobo Xia, Shiming Ge, and Tongliang Liu.
\newblock Selective-supervised contrastive learning with noisy labels.
\newblock In {\em CVPR}, 2022.

\bibitem{DBLP:journals/tsmc/LiuWZ09}
Xu{-}Ying Liu, Jianxin Wu, and Zhi{-}Hua Zhou.
\newblock Exploratory undersampling for class-imbalance learning.
\newblock {\em {IEEE} Trans. Syst. Man Cybern. Part {B}}, 2009.

\bibitem{DBLP:conf/icml/LukasikBMK20}
Michal Lukasik, Srinadh Bhojanapalli, Aditya~Krishna Menon, and Sanjiv Kumar.
\newblock Does label smoothing mitigate label noise?
\newblock In {\em ICML}, 2020.

\bibitem{DBLP:conf/iclr/MenonJRJVK21}
Aditya~Krishna Menon, Sadeep Jayasumana, Ankit~Singh Rawat, Himanshu Jain,
  Andreas Veit, and Sanjiv Kumar.
\newblock Long-tail learning via logit adjustment.
\newblock In {\em ICLR}, 2021.

\bibitem{DBLP:conf/cvpr/OrtegoAAOM21}
Diego Ortego, Eric Arazo, Paul Albert, Noel~E. O'Connor, and Kevin McGuinness.
\newblock Multi-objective interpolation training for robustness to label noise.
\newblock In {\em CVPR}, 2021.

\bibitem{Park_2021_ICCV}
Seulki Park, Jongin Lim, Younghan Jeon, and Jin~Young Choi.
\newblock Influence-balanced loss for imbalanced visual classification.
\newblock In {\em ICCV}, 2021.

\bibitem{DBLP:conf/cvpr/PatriniRMNQ17}
Giorgio Patrini, Alessandro Rozza, Aditya~Krishna Menon, Richard Nock, and
  Lizhen Qu.
\newblock Making deep neural networks robust to label noise: {A} loss
  correction approach.
\newblock In {\em CVPR}, 2017.

\bibitem{DBLP:conf/iclr/PereyraTCKH17}
Gabriel Pereyra, George Tucker, Jan Chorowski, Lukasz Kaiser, and Geoffrey~E.
  Hinton.
\newblock Regularizing neural networks by penalizing confident output
  distributions.
\newblock In {\em ICLR}, 2017.

\bibitem{REED200115}
William~J Reed.
\newblock The pareto, zipf and other power laws.
\newblock {\em Economics Letters}, 74(1):15--19, 2001.

\bibitem{DBLP:conf/nips/RenYSMZYL20}
Jiawei Ren, Cunjun Yu, Shunan Sheng, Xiao Ma, Haiyu Zhao, Shuai Yi, and
  Hongsheng Li.
\newblock Balanced meta-softmax for long-tailed visual recognition.
\newblock In {\em NeurIPS}, 2020.

\bibitem{DBLP:conf/nips/RenHGS15}
Shaoqing Ren, Kaiming He, Ross~B. Girshick, and Jian Sun.
\newblock Faster {R-CNN:} towards real-time object detection with region
  proposal networks.
\newblock In {\em NeurIPS}, 2015.

\bibitem{DBLP:conf/nips/ShuXY0ZXM19}
Jun Shu, Qi Xie, Lixuan Yi, Qian Zhao, Sanping Zhou, Zongben Xu, and Deyu Meng.
\newblock Meta-weight-net: Learning an explicit mapping for sample weighting.
\newblock In {\em NeurIPS}, 2019.

\bibitem{DBLP:conf/nips/SohnBCZZRCKL20}
Kihyuk Sohn, David Berthelot, Nicholas Carlini, Zizhao Zhang, Han Zhang, Colin
  Raffel, Ekin~Dogus Cubuk, Alexey Kurakin, and Chun{-}Liang Li.
\newblock Fixmatch: Simplifying semi-supervised learning with consistency and
  confidence.
\newblock In {\em NeurIPS}, 2020.

\bibitem{DBLP:conf/icml/SongK019}
Hwanjun Song, Minseok Kim, and Jae{-}Gil Lee.
\newblock {SELFIE:} refurbishing unclean samples for robust deep learning.
\newblock In {\em ICML}, 2019.

\bibitem{DBLP:journals/corr/abs-2007-08199}
Hwanjun Song, Minseok Kim, Dongmin Park, and Jae{-}Gil Lee.
\newblock Learning from noisy labels with deep neural networks: {A} survey.
\newblock {\em CoRR}, 2020.

\bibitem{DBLP:conf/cvpr/TanWLLOYY20}
Jingru Tan, Changbao Wang, Buyu Li, Quanquan Li, Wanli Ouyang, Changqing Yin,
  and Junjie Yan.
\newblock Equalization loss for long-tailed object recognition.
\newblock In {\em CVPR}, 2020.

\bibitem{DBLP:conf/iclr/WangLM0Y21}
Xudong Wang, Long Lian, Zhongqi Miao, Ziwei Liu, and Stella~X. Yu.
\newblock Long-tailed recognition by routing diverse distribution-aware
  experts.
\newblock In {\em ICLR}, 2021.

\bibitem{DBLP:conf/cvpr/WeiSMYY21}
Chen Wei, Kihyuk Sohn, Clayton Mellina, Alan~L. Yuille, and Fan Yang.
\newblock Crest: {A} class-rebalancing self-training framework for imbalanced
  semi-supervised learning.
\newblock In {\em CVPR}, 2021.

\bibitem{wei2022learning}
Jiaheng Wei, Zhaowei Zhu, Hao Cheng, Tongliang Liu, Gang Niu, and Yang Liu.
\newblock Learning with noisy labels revisited: A study using real-world human
  annotations.
\newblock In {\em ICLR}, 2022.

\bibitem{DBLP:journals/corr/abs-2108-11569}
Tong Wei, Jiang{-}Xin Shi, Wei{-}Wei Tu, and Yu{-}Feng Li.
\newblock Robust long-tailed learning under label noise.
\newblock {\em CoRR}, 2021.

\bibitem{DBLP:conf/iclr/XiaL00WGC21}
Xiaobo Xia, Tongliang Liu, Bo Han, Chen Gong, Nannan Wang, Zongyuan Ge, and Yi
  Chang.
\newblock Robust early-learning: Hindering the memorization of noisy labels.
\newblock In {\em ICLR}, 2021.

\bibitem{xia2022sample}
Xiaobo Xia, Tongliang Liu, Bo Han, Mingming Gong, Jun Yu, Gang Niu, and Masashi
  Sugiyama.
\newblock Sample selection with uncertainty of losses for learning with noisy
  labels.
\newblock In {\em ICLR}, 2022.

\bibitem{DBLP:conf/cvpr/XiaoXYHW15}
Tong Xiao, Tian Xia, Yi Yang, Chang Huang, and Xiaogang Wang.
\newblock Learning from massive noisy labeled data for image classification.
\newblock In {\em CVPR}, 2015.

\bibitem{yi2022identifying}
Yi Xuanyu, Tang Kaihua, Hua Xian-Sheng, Lim Joo-Hwee, and Zhang Hanwang.
\newblock Identifying hard noise in long-tailed sample distribution.
\newblock In {\em ECCV}, 2022.

\bibitem{DBLP:conf/nips/YaoL0GD0S20}
Yu Yao, Tongliang Liu, Bo Han, Mingming Gong, Jiankang Deng, Gang Niu, and
  Masashi Sugiyama.
\newblock Dual {T:} reducing estimation error for transition matrix in
  label-noise learning.
\newblock In {\em NeurIPS}, 2020.

\bibitem{DBLP:conf/cvpr/YaoSZS00T21}
Yazhou Yao, Zeren Sun, Chuanyi Zhang, Fumin Shen, Qi Wu, Jian Zhang, and
  Zhenmin Tang.
\newblock Jo-src: {A} contrastive approach for combating noisy labels.
\newblock In {\em CVPR}, 2021.

\bibitem{DBLP:conf/cvpr/00010S0C19}
Xi Yin, Xiang Yu, Kihyuk Sohn, Xiaoming Liu, and Manmohan Chandraker.
\newblock Feature transfer learning for face recognition with under-represented
  data.
\newblock In {\em CVPR}, 2019.

\bibitem{DBLP:conf/iccv/ZangHL21}
Yuhang Zang, Chen Huang, and Chen~Change Loy.
\newblock {FASA:} feature augmentation and sampling adaptation for long-tailed
  instance segmentation.
\newblock In {\em ICCV}, 2021.

\bibitem{DBLP:conf/iclr/ZhangCDL18}
Hongyi Zhang, Moustapha Ciss{\'{e}}, Yann~N. Dauphin, and David Lopez{-}Paz.
\newblock mixup: Beyond empirical risk minimization.
\newblock In {\em ICLR}, 2018.

\bibitem{DBLP:conf/cvpr/ZhangLY0S21}
Songyang Zhang, Zeming Li, Shipeng Yan, Xuming He, and Jian Sun.
\newblock Distribution alignment: {A} unified framework for long-tail visual
  recognition.
\newblock In {\em CVPR}, 2021.

\bibitem{DBLP:conf/aaai/ZhangWZW21}
Yongshun Zhang, Xiu{-}Shen Wei, Boyan Zhou, and Jianxin Wu.
\newblock Bag of tricks for long-tailed visual recognition with deep
  convolutional neural networks.
\newblock In {\em AAAI}, 2021.

\bibitem{DBLP:conf/cvpr/ZhongC0J21}
Zhisheng Zhong, Jiequan Cui, Shu Liu, and Jiaya Jia.
\newblock Improving calibration for long-tailed recognition.
\newblock In {\em CVPR}, 2021.

\bibitem{DBLP:conf/aaai/ZhuN0Z22}
Beier Zhu, Yulei Niu, Xian{-}Sheng Hua, and Hanwang Zhang.
\newblock Cross-domain empirical risk minimization for unbiased long-tailed
  classification.
\newblock In {\em AAAI}, 2022.

\end{thebibliography}
}

\newpage
\appendix
\onecolumn

\begin{center} 
	{\centering\section*{Appendix}}
\end{center}
\setcounter{table}{0}   
\setcounter{figure}{0}
\renewcommand{\thetable}{A\arabic{table}}
\renewcommand{\thefigure}{A\arabic{figure}}
%
\section{Pseudocode of the Proposed Method}\label{pseudocode}
Algorithm \ref{alg:algorithm} details the training procedure of the proposed framework for label-noise learning with intrinsically long-tailed data.

\begin{algorithm}[!h]\label{algorithm}
	\caption{The training process of the proposed learning framework}
	\label{alg:algorithm}
	\LinesNumbered 
	\KwIn{Noisy training data $\DD=\{x_i,\hat{y}_i\}_{i=1}^N$}
	Initialize the model $\theta$ trained on $\DD$;\\
	\While{$e<\mathrm{MaxIterationNumber}$}{
		\For{$c=1$ \KwTo $M$}{
            \For{$i=1$ \KwTo $|\DD_c|$}{
                Obtain and store the feature $\mathbf{f}_i$ and the prediction confidence $\mathbf{p}_i$ for $x_i$ by model $\theta$;\\
            }
            Calculate the average prediction confidence $\bar{\mathbf{p}}_c$ for class $c$;\\
		}
	   \For{$c=1$ \KwTo $M$}{
            Calculate the confidence thresholds $H_c$  for class $c$ by Equation (\ref{hlc});\\
            Calculate the adaptive centroid $\mathbf{o}_c$ for class $c$ by Equation (\ref{adc});\\
            \For{$i=1$ \KwTo $|\DD_c|$}{
                Calculate the additional weight $W(x_i)$ by Equation (\ref{w});\\
                Calculate $WJSD(x_i)$ by Equation (\ref{wjsd});\\
                Calculate $ACD(x_i)$ by Equation (\ref{acd});\\
            }
            Apply GMM with values of bi-dimensional metrics to all the samples in class $c$;\\
            Adopt dimension selection for class $c$ by Algorithnm \ref{alg:dimension_selection};\\
            Adopt cluster selection for selected dimension by Equation (\ref{cluster_selection_}) to generate two sets $\DD_c^{clean}$ and $\DD_c^{noisy}$;\\
		}
		Update the model $\theta$ by SSL training with $\DD^{clean}$ as labeled data and $\DD^{noisy}$ as unlabeled data.
	}
\end{algorithm}
\section{Proof of Theorem 1}\label{proof}
\noindent\textit{Proof.}
Suppose $x_i$ and $x_j$ are two samples in class $c$, $p_i^c$ and $q_i^c$ are the $c$'s dimension of their prediction confidence $\mathbf{p}_i=[p_i^1,p_i^2,...,p_i^M]$ and $\mathbf{p}_j=[p_j^1,p_j^2,...,p_j^M]$, respectively.
Their common observed class label $\hat{\mathbf{y}}=[\hat{y}_j^1,\hat{y}_j^2,...,\hat{y}_j^M]$ is in the one-hot form where only the value of the $c$'s dimension is 1 ($\hat{y}_i^c=1$), and the values on other dimensions are all 0 ($\hat{y}_i^d=0$ for all $d\ne c$). Jensen-Shannon Divergence (JSD) for sample $x_i$ is defined as:
\begin{align}\label{jsd}
JSD(x_i)
=&\frac{1}{2} KL\Big(\mathbf{p}_i \Big\| \frac{\mathbf{p}_i +{\hat{\mathbf{y}}_i}}{2}\Big)+\frac{1}{2} KL\Big({\hat{\mathbf{y}_i}}  \Big\| \frac{\mathbf{p}_i +{\hat{\mathbf{y}}_i}}{2}\Big)\\
=&\frac{1}{2}\left(\sum_{d=1}^Mp_i^d\log\frac{2p_i^d}{p_i^d+\hat{y}_i^d}+\sum_{d=1}^M\hat{y}_i^d\log\frac{2\hat{y}_i^d}{p_i^d+\hat{y}_i^d}\right)\\
=&\frac{1}{2}\left(\sum_{d\ne c}p_i^d\log\frac{2p_i^d}{p_i^d}+p_i^c\log\frac{2p_i^c}{p_i^c+\hat{y}_i^c}+\hat{y}_i^c\log\frac{2\hat{y}_i^c}{p_i^c+\hat{y}_i^c}\right)\\
=&\frac{1}{2}\left(\sum_{d\ne c}p_i^d+p_i^c\log\frac{2p_i^c}{p_i^c+\hat{y}_i^c}+\log\frac{2}{p_i^c+1}\right)\\
\end{align}
\begin{align}\label{jsd}
\quad\quad\quad\quad\quad\quad\quad\quad  =&\frac{1}{2}\left(1-p_i^c+p_i^c\log\frac{2p_i^c}{p_i^c+1}+\log\frac{2}{p_i^c+1}\right)\\
=&\frac{1}{2}\left(1-p_i^c+p_i^c+p_i^c\log{p_i^c}-p_i^c\log{(p_i^c+1)}+1-\log{(p_i^c+1)}\right)\\
=&\frac{1}{2}\left(2+p_i^c\log{p_i^c}-(p_i^c+1)\log{(p_i^c+1)}\right).
\end{align}
The base of logarithm is 2 for the above derivation. 
Then
\begin{align}\label{}
|JSD(x_i)-JSD(x_j)|&=\left|\frac{1}{2}\left(2+p_i^c\log{p_i^c}-(p_i^c+1)\log{(p_i^c+1)}\right)-\frac{1}{2}\left(2+p_j^c\log{p_j^c}-(p_j^c+1)\log{(p_j^c+1)}\right)\right|\\
&=\frac{1}{2}\left|(p_i^c\log{p_i^c}-(p_i^c+1)\log{(p_i^c+1)}\right)-\left(p_j^c\log{p_j^c}-(p_j^c+1)\log{(p_j^c+1)}\right|
\end{align}
Now, for $u\in(0,1)$, let
\begin{align}\label{}
f(u)=u\log u-(u+1)\log(u+1).
\end{align}
The first- and second-order derivative of $f(u)$ can be obtained by:
\begin{align}\label{}
f'(u)&=\left(\log u+u\cdot\frac{1}{u}\cdot\frac{1}{\ln 2}\right) - \left(\log (u+1)+(u+1)\cdot\frac{1}{u+1}\cdot\frac{1}{\ln 2}\right)\\
&=\log u-\log (u+1).\\
f''(u)&=\left(\frac{1}{u}-\frac{1}{u+1}\right)\frac{1}{\ln 2}.
\end{align}
It can be observed that $f'(u)<0$ and $f''(u)>0$ that makes $f'(u)$ monotonically increase and $|f'(u)|$ monotonically decrease. By Lagrange mean value theorem, if function $f(u)$ is continuous and differentiable on the interval $(0,1)$, then there is at least one point $\xi$ between two real numbers $a,b\in(0,1)$:
\begin{align}\label{}
|f(a)-f(b)|=|f'(\xi)(a-b)|=|f'(\xi)|\cdot|a-b|.
\end{align}
As $\xi>a$ and $|f'(u)|$ monotonically decrease, we have:
\begin{align}\label{}
|f'(\xi)|\cdot|a-b|\le&|f'(a)|\cdot|a-b|\\
=&|\log a-\log (a+1)|\cdot|a-b|\\
=&\left|\log \frac{a}{a+1}\right|\cdot|a-b|\\
=&\log \frac{a+1}{a}\cdot|a-b|
\end{align}
By replacing $f(u)$ by $JSD(x)$, $a$ by $p_i^c$ and $b$ by $p_j^c$, we have:
\begin{align}\label{}
|JSD(x_i)-JSD(x_j)|\le \frac{1}{2}\log\left(\frac{p_i^c+1}{p_i^c}\right)|p_i^c-p_j^c|.
\end{align}

\section{The Correlation between Purity and Clean Ratio}\label{correlation_p}
\begin{figure}[!h]
\centering
    \includegraphics[width=0.4\textwidth]{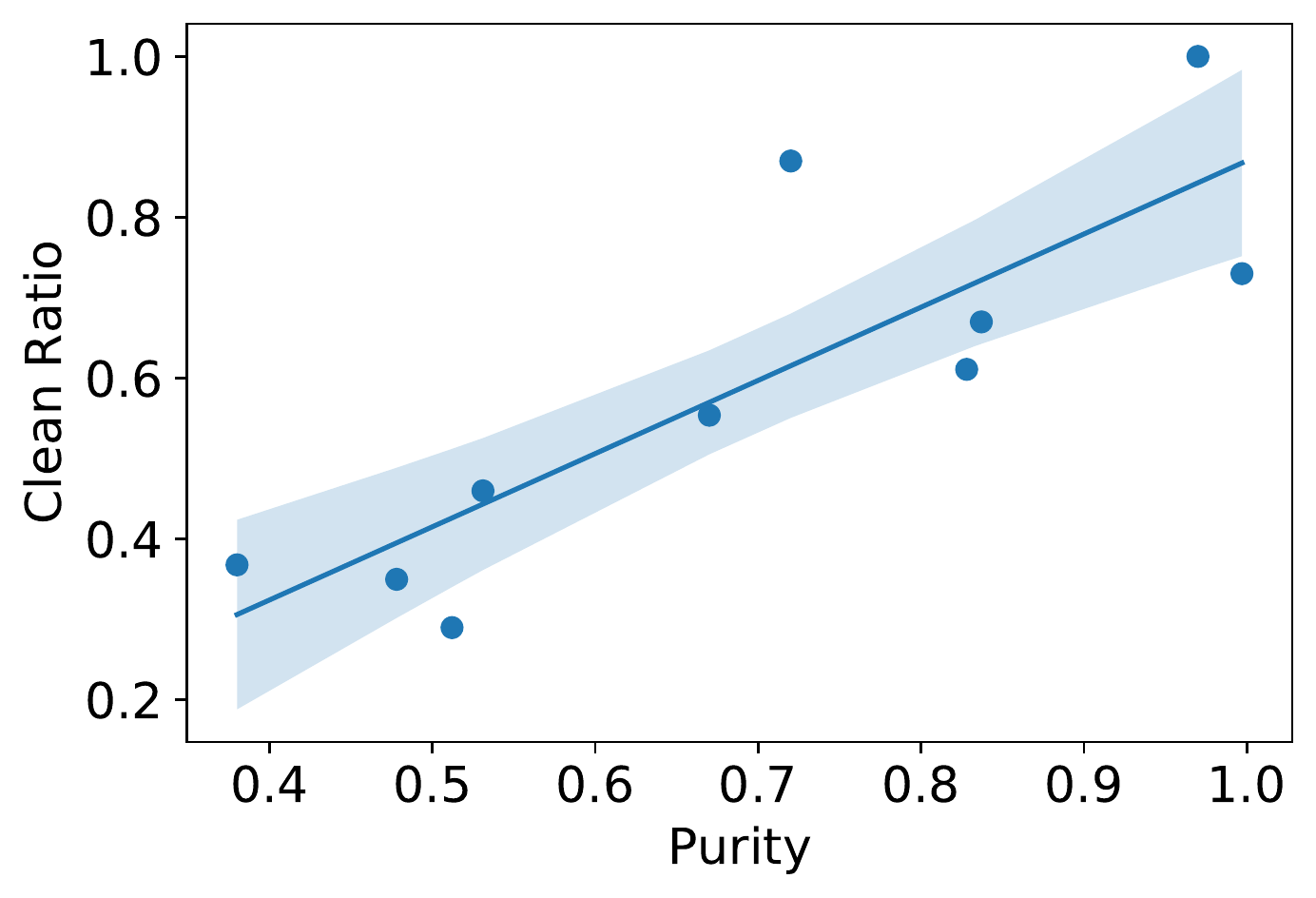}
\caption{The correlation between purity and clean ratio.}
\label{purity_correlation}
\end{figure}

To validate the promotion effect of purity on the separation of clean and noisy samples, we show the highest proportion of clean samples for clusters obtained with different purity in the tail class with asymmetric noise.
\cref{purity_correlation} describes the relationship between purity and the proportion of clean samples in the cluster. 
It can be observed that there is a positive correlation between purity and the proportion of clean samples in the cluster.
Therefore, enhancing the purity of centroid can effectively improve the separation ability of clean and noisy samples.

\section{The Purity by High-Confidence Sample Set}\label{high_p}
\begin{figure}[!h]
\centering
    \includegraphics[width=0.4\textwidth]{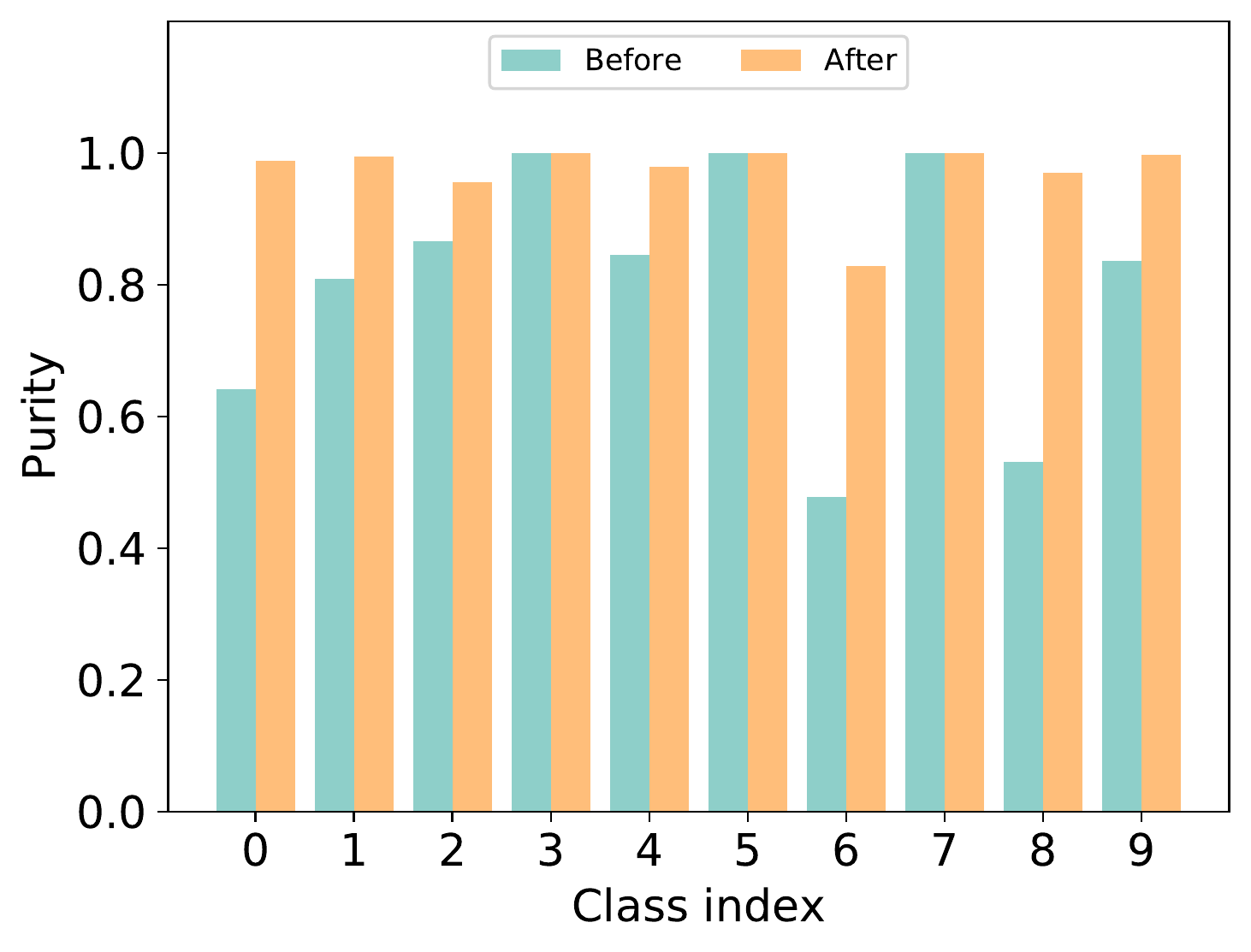}
\caption{The purity difference before and after the selection of high confidence samples.}
\label{purity_effect}
\end{figure}

In order to validate the effectiveness of our proposed high-confidence sample set in terms of purity improvement, we compared the purity difference before and after the selection of high-confidence samples in CIFAR-10 with 0.4 asymmetric noise and 0.01 imbalance factor.
As shown in \cref{purity_effect}, the purity of all classes is significantly improved through the selection of high-confidence samples, and the improvement of the tail class is more prominent.
It illustrates the robustness of our approach to the long-tail distribution.

\begin{table}[!h]
        \centering
        \begin{tabular}{@{}lllllcllll@{}}
                \toprule
                Dataset       & \multicolumn{2}{c}{CIFAR-10}      & \multicolumn{2}{c}{CIFAR-100}       \\ 
                \midrule
                                Noise Ratio  & 
                \multicolumn{4}{c}{0.4}  \\ \midrule
                Imbalance Factor       & \multicolumn{1}{c}{0.1} & \multicolumn{1}{c}{0.01}  & \multicolumn{1}{c}{0.1} & \multicolumn{1}{c}{0.01}   \\
                \midrule

                baseline-clean  & {87.88} & {70.11} & {59.75} & {42.65}\\
                cRT-full  & {84.62} & {74.53} & {55.94} & {46.22}\\
                 \bottomrule
        \end{tabular}
        \caption{Performance comparison with asymmetric noise and long-tail distribution.}
        \label{feature_effct}
\end{table}

\section{The Effectiveness of Feature in Asymmetric Noise}
In order to show why the feature  is good enough to discriminate the sample under asymmetric noise, we retrain the classifier using full clean samples (cRT-full) after fixing the features for the model trained under asymmetric noise with long-tail distribution.
Accordingly, we also use clean and long-tail distribution samples to train the model as a baseline for comparison.
\cref{feature_effct} reports the accuracy of both approaches.
It can be observed that the potential of the model under the influence of noise and long-tail distribution is similar to the performance of the model trained directly under the clean and long-tail distribution samples.
It means that the model can still learn good features to distinguish between samples even under the dual effects of the noise and long-tail distribution.
It is also an important guarantee to use features to separate samples under long-tail distribution and asymmetric noise.

\section{More Experimental Results on CIFAR-10/100}
In this section, we compare the performance of different methods with symmetric and asymmetric noise in 0.01 imbalance factor, which is a more difficult scenario.
\cref{cifar_noise} reports the accuracy of different methods in these settings.
One of the results of RoLT \cite{DBLP:journals/corr/abs-2108-11569} is empty because the code does not run properly in this case.
It can be observed that our proposed method achieves the best performance in most cases, which further verifies the effectiveness of our method.


\begin{table}[!t] 
   \begin{subtable}{0.5\columnwidth}
        \centering
        \begin{tabular}{@{}l|llllcllll@{}}
                \toprule
                \multicolumn{2}{l}{Dataset}       & \multicolumn{2}{c}{CIFAR-10}      & \multicolumn{2}{c}{CIFAR-100}       \\ 
                \midrule
                \multicolumn{2}{l}{Imbalance Factor}       & \multicolumn{4}{c}{0.01}     \\
                \midrule
                \multicolumn{2}{l}{Noise Ratio (\textbf{Sym.})}  & 
                \multicolumn{1}{c}{0.4} & \multicolumn{1}{c}{0.6} & \multicolumn{1}{c}{0.4} & \multicolumn{1}{c}{0.6}\\ \midrule
                Baseline & CE & {47.81} & {28.04} & {21.99} & {15.51}\\ \midrule
                 \multirow{3}{*}{LT}& LA   & {42.63} & {36.37}   & {21.54} & {13.14}    \\
                 & LDAM  & {45.52} & {35.29}   & {18.81} & {12.65}    \\
                 & IB  & {49.07} & {32.54}   & {20.34} & {12.10}    \\  \midrule
               \multirow{2}{*}{NL} &DivideMix    & {32.42} & {34.73} & {36.20} & {\textbf{26.29}}\\
                &UNICON      & {61.23} & {54.69} & {32.09} & {24.82}\\\midrule
               \multirow{4}{*}{NL-LT} &MW-Net  & {46.62} & {39.33} & {19.65} & {13.72}\\
                &RoLT      & {60.11} & {44.23} & {23.51} & {16.61}\\
                &HAR       & {51.54} & {38.28} & {20.21} & {14.89} \\
                &ULC       & {45.22} & {50.56} & {33.41} & {25.69}\\\midrule
                Our&TABASCO    & {\textbf{62.34}} & {\textbf{55.76}} & {\textbf{36.91}} & {26.25}\\ 
                \bottomrule
        \end{tabular}

        \end{subtable} %
\begin{subtable}{0.5\columnwidth}
        \centering
        \begin{tabular}{@{}l|llllcllll@{}}
                \toprule
                \multicolumn{2}{l}{Dataset}       & \multicolumn{2}{c}{CIFAR-10}      & \multicolumn{2}{c}{CIFAR-100}       \\ 
                \midrule
                \multicolumn{2}{l}{Imbalance Factor}       & \multicolumn{4}{c}{0.01}     \\
                \midrule
                \multicolumn{2}{l}{Noise Ratio (\textbf{Asym.})}  & 
                \multicolumn{1}{c}{0.2} & \multicolumn{1}{c}{0.4} & \multicolumn{1}{c}{0.2} & \multicolumn{1}{c}{0.4}\\ \midrule
                Baseline & CE & {56.56} & {44.64} & {25.35} & {17.89}\\ \midrule
                 \multirow{3}{*}{LT}& LA   & {58.78} & {43.37}   & {32.16} & {22.67}    \\
                 & LDAM  & {61.25} & {40.85}   & {29.22} & {18.65}    \\
                 & IB  & {56.28} & {42.96}   & {31.15} & {23.40}    \\  \midrule
               \multirow{2}{*}{NL} &DivideMix    & {41.12} & {42.79} & {38.46} & {29.69}\\
                &UNICON      & {53.53} & {34.05} & {34.14} & {30.72}\\\midrule
               \multirow{4}{*}{NL-LT} &MW-Net  & {62.19} & {45.21} & {27.56} & {20.04}\\
                &RoLT      & {54.81} & {50.26} & {32.96} & \makecell[c]{-}\\
                &HAR       & {62.42} & {51.97} & {27.90} & {20.03} \\
                &ULC       & {41.14} & {22.73} & {34.07} & {25.04}\\\midrule
                Our&TABASCO    & {\textbf{62.98}} & {\textbf{54.04}} & {\textbf{40.35}} & {\textbf{33.15}}\\ 
                \bottomrule
        \end{tabular}
        \end{subtable}
                \caption{Performance comparison with synthetic noise and long-tail distribution.
The best results are shown in bold.}
\label{cifar_noise}
\end{table}

\begin{table}[!h]
        \centering
        \resizebox{0.45\columnwidth}{!}{
        \begin{tabular}{@{}l|llllcllll@{}}
                \toprule
                \multicolumn{2}{l}{Dataset}  & \multicolumn{2}{c}{Red}   & \multicolumn{1}{c}{10N} & \multicolumn{1}{c}{100N}      \\ 
                \midrule
                \multicolumn{2}{l}{Imbalance Factor}    & \multicolumn{2}{c}{$\approx$ 0.01}   & \multicolumn{2}{c}{0.01}     \\
                \midrule
                \multicolumn{2}{l}{Noise Ratio}  & 
                \multicolumn{1}{c}{0.2} & \multicolumn{1}{c}{0.4} & \multicolumn{2}{c}{$\approx$ 0.4} \\ \midrule
                Baseline & CE & {30.88} & {31.46} & {49.31} & {25.28}\\ \midrule
                 \multirow{3}{*}{LT}& LA   & {10.32} & {9.560}   & {50.09} & {26.39}    \\
                 & LDAM  & {14.30} & {15.64}   & {50.36} & {30.17}    \\
                 & IB  & {16.72} & {16.34}   & {56.41} & {31.55}    \\  \midrule
                 \multirow{2}{*}{NL}&  DivideMix    & {33.00} & {34.72} & {30.67} & {31.34}\\
                 & UNICON      & {31.86} & {31.12} & {59.47} & {37.06}\\\midrule
                \multirow{4}{*}{NL-LT}&  MW-Net  & {30.74} & {31.12} & {54.95} & {31.80}\\
                 &RoLT      & {15.78} & {16.90} & {61.23} & {33.48}\\
                &  HAR       & {32.60} & {31.30} & {56.84} & {32.34}\\
                &ULC      & {34.24} & {34.84}    & {43.89} & {35.71} \\  \midrule
                 \multirow{1}{*}{Our}& TABASCO   & {\textbf{37.20}} & {\textbf{37.12}} & {\textbf{64.54}} & {\textbf{39.30}}\\ 
                 \bottomrule
        \end{tabular}
        }
        \caption{Performance comparison with real-world noise and long-tail distribution.
The best results are shown in bold.}
        \label{realist_datasets_2}
\end{table}
\section{More Experimental Results on Benchmarks}
In this section, we compare the performance of different methods with realist noise in 0.01 imbalance factor, which is a more difficult scenario.
\cref{realist_datasets_2} reports the accuracy of different methods in these settings.
It can be observed that our proposed method achieves the best performance in all cases, which further verifies the effectiveness of our method.

\section{The Observed  and Intrinsic Distribution of Benchmarks}
\begin{figure}[!h]
\begin{subfigure}[b]{0.33\columnwidth}
    \includegraphics[width=\textwidth]{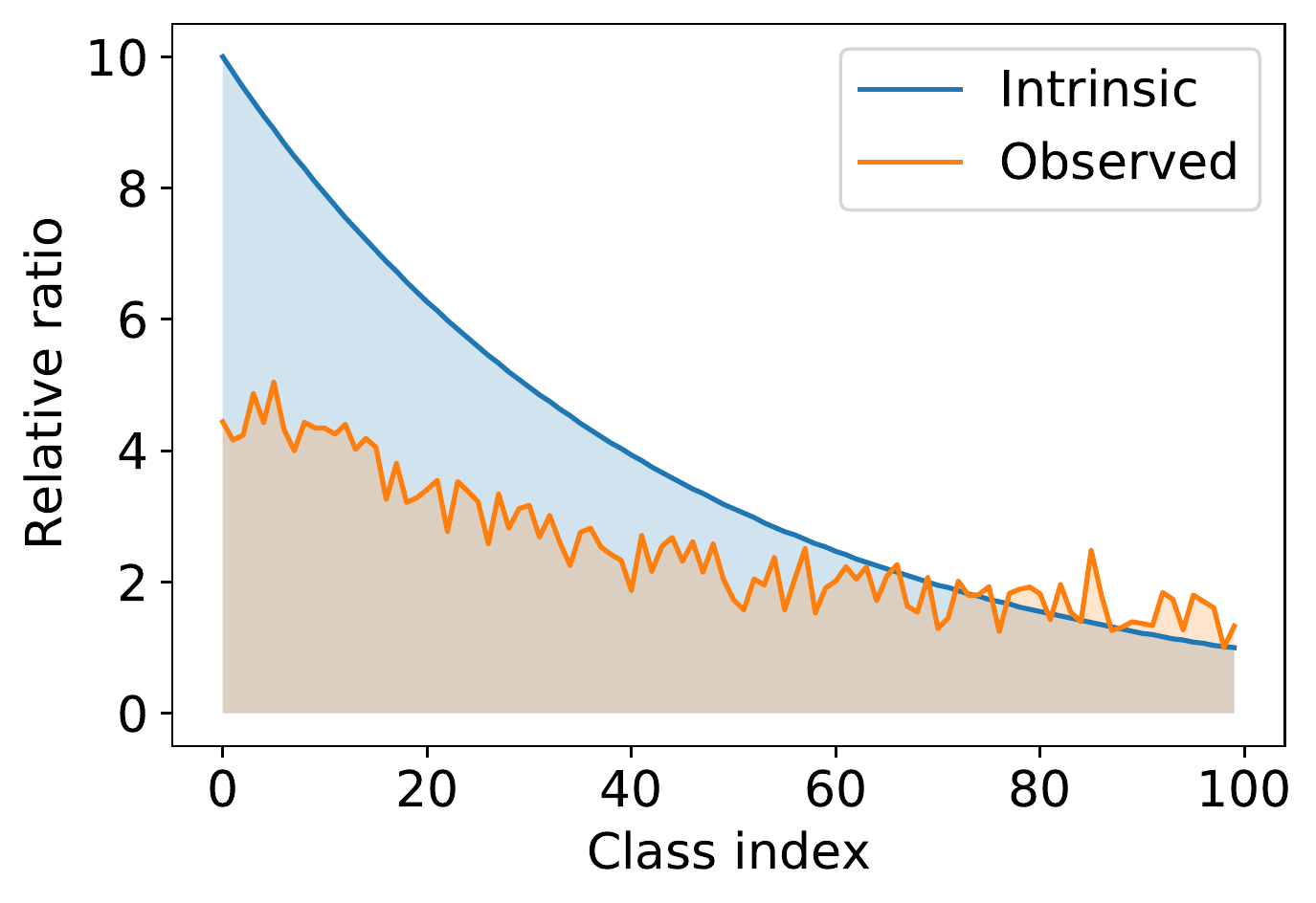}

    \subcaption{Red}
\end{subfigure} 
\begin{subfigure}[b]{0.33\columnwidth}
    \includegraphics[width=\textwidth]{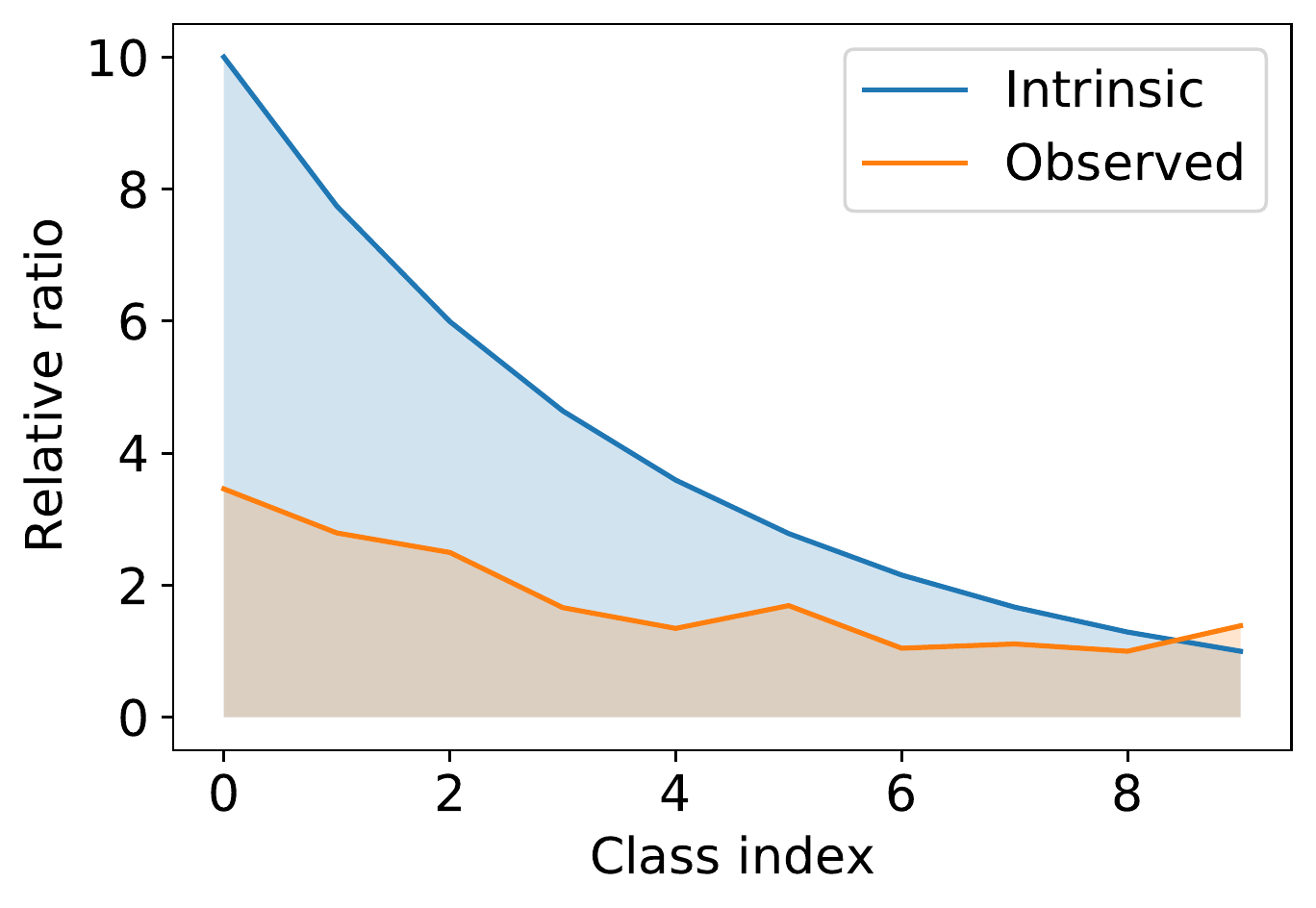}

    \subcaption{10N}
\end{subfigure}
\begin{subfigure}[b]{0.33\columnwidth}
    \includegraphics[width=\textwidth]{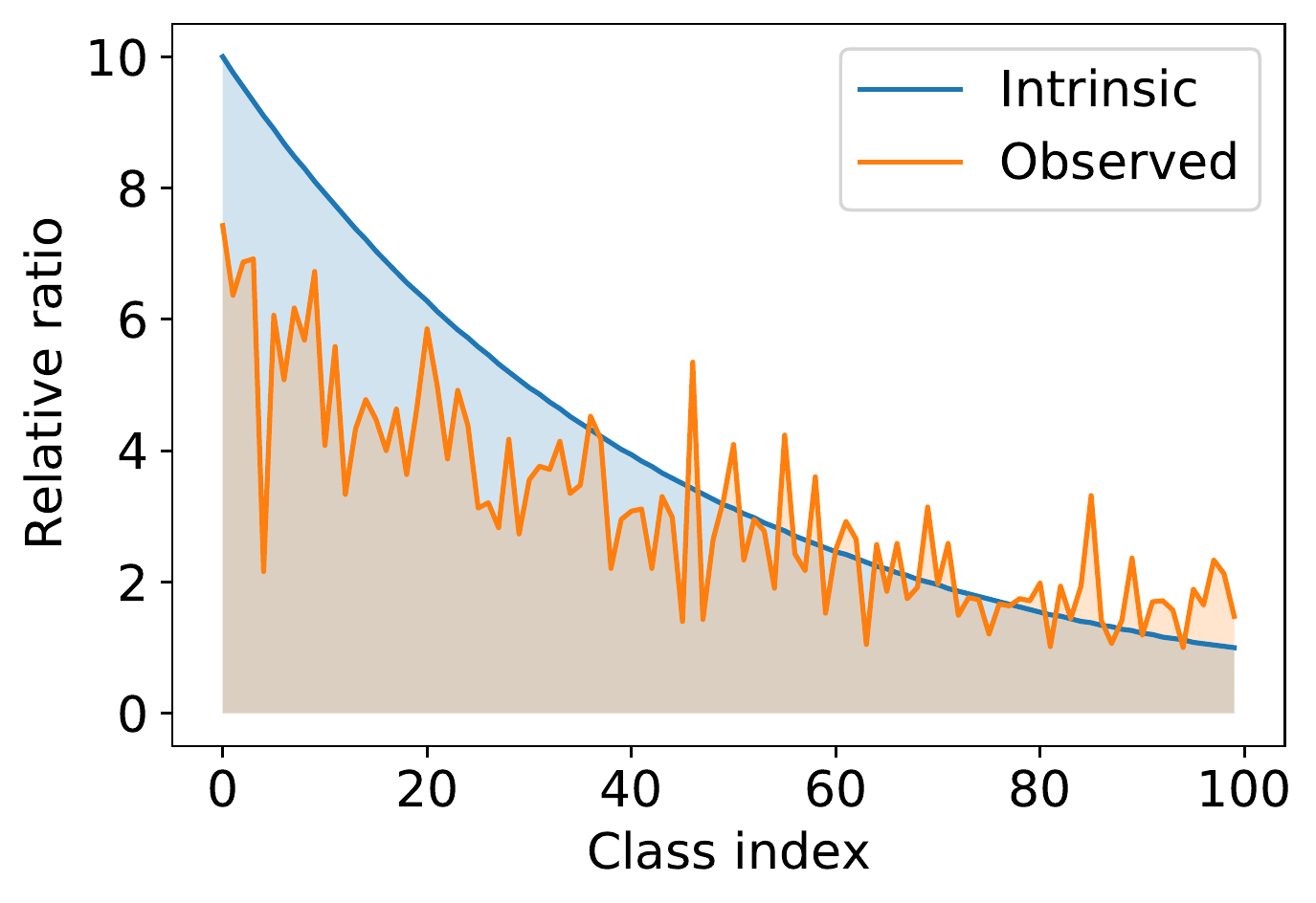}

    \subcaption{100N}
\end{subfigure}
\caption{The observed  and intrinsic distribution of benchmarks.}
\label{dis_app}
\end{figure}

In this section, we plot the observed  and  intrinsic distribution of benchmarks we proposed in \cref{dis_app}.
We use benchmarks with an imbalance factor of approximately 0.1  and a noise ratio of approximately 0.4 for our analysis.
Relative ratio is denoted by the ratio between the number of samples in each class and the minimum number of samples.
It can be observed that the intrinsic and observed distributions of all benchmarks are significantly different, and observed distribution is more balanced.
This means that it is difficult to focus on  the intrinsic tail class and the tail class has a high percentage of noise, making it more difficult to separate noise and clean samples in the intrinsic tail class.
It also corresponds to the two key challenges of the problem  we present and demonstrates that our proposed benchmarks are a good measure of the effectiveness of different approaches to the problem.
\section{More Sample Distribution on Bi-dimensional Separation Metric}
\begin{figure}[!h]
\hspace{8px}
\begin{subfigure}[b]{0.5\columnwidth}
    \includegraphics[width=0.45\textwidth]{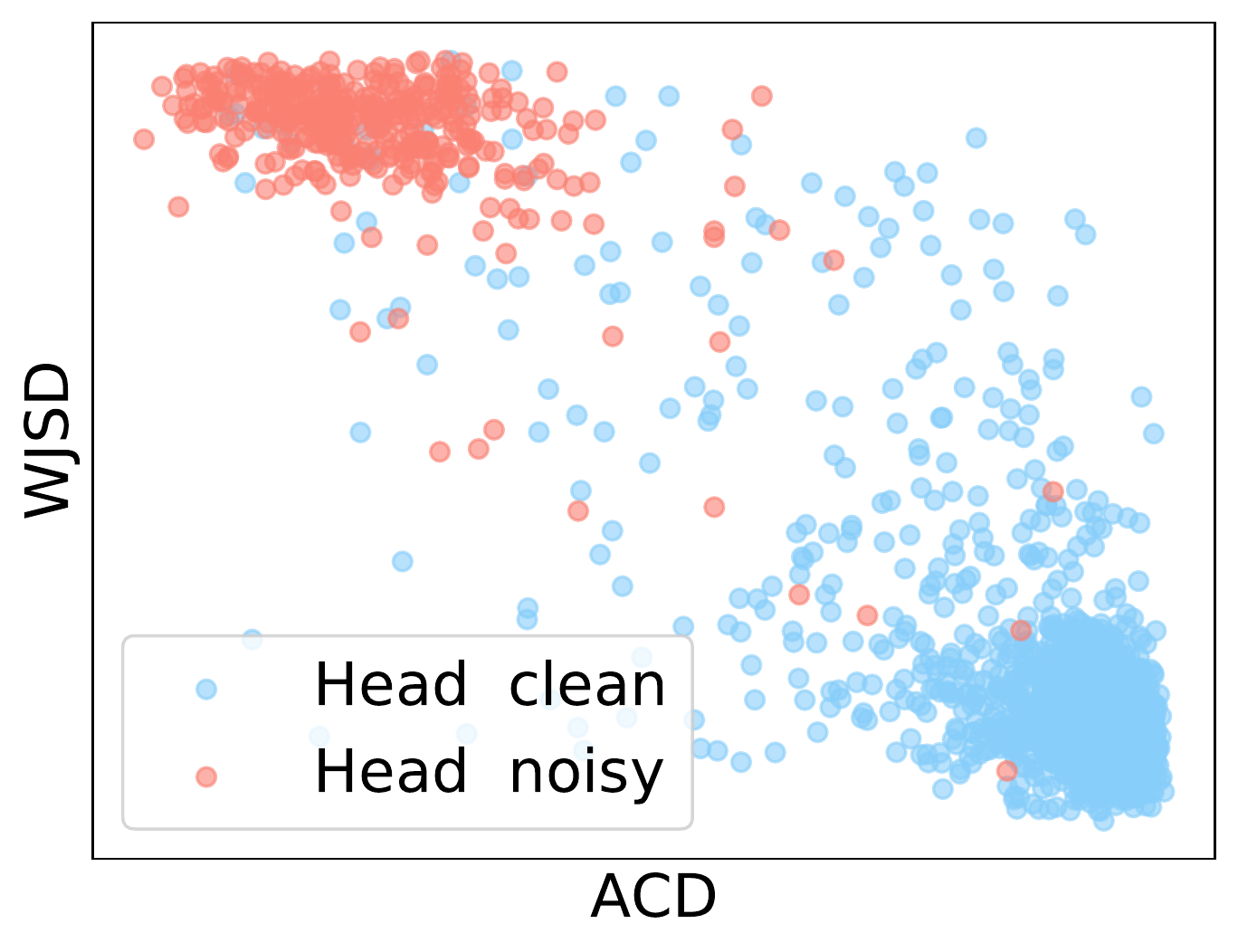}
    \includegraphics[width=0.45\textwidth]{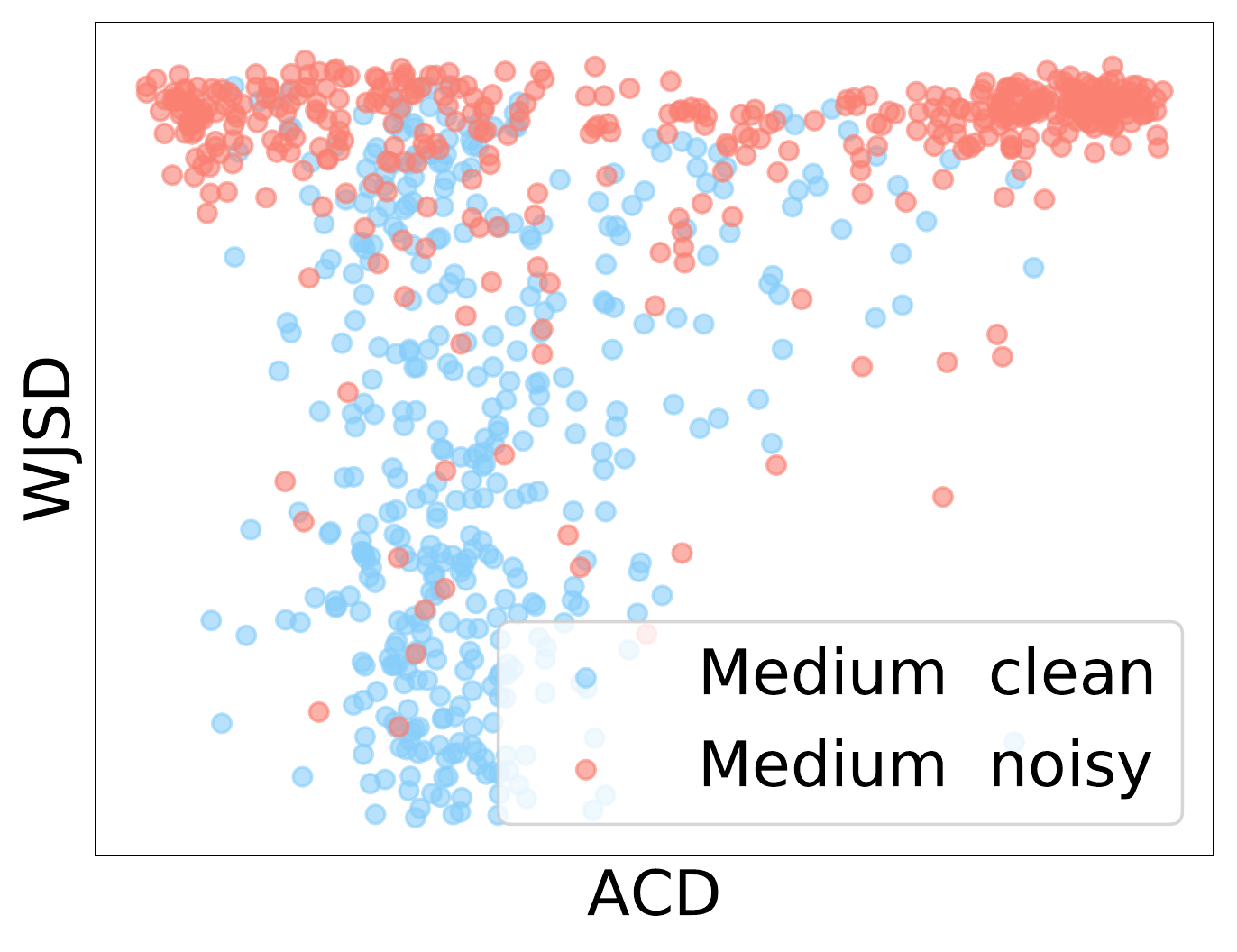}
    \subcaption{Symmetric noise}
\end{subfigure} 
\begin{subfigure}[b]{0.5\columnwidth}
    \includegraphics[width=0.45\textwidth]{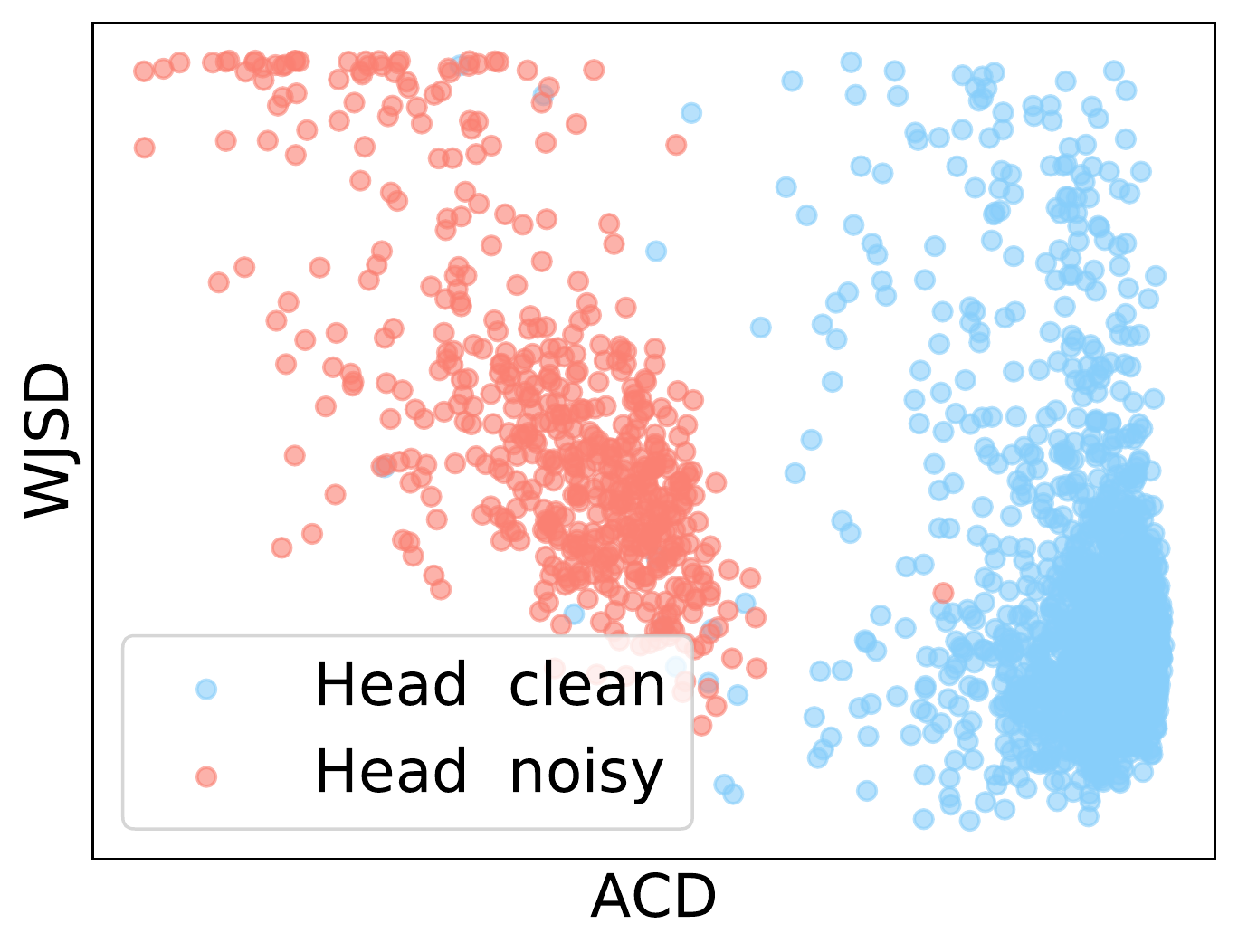}
    \includegraphics[width=0.45\textwidth]{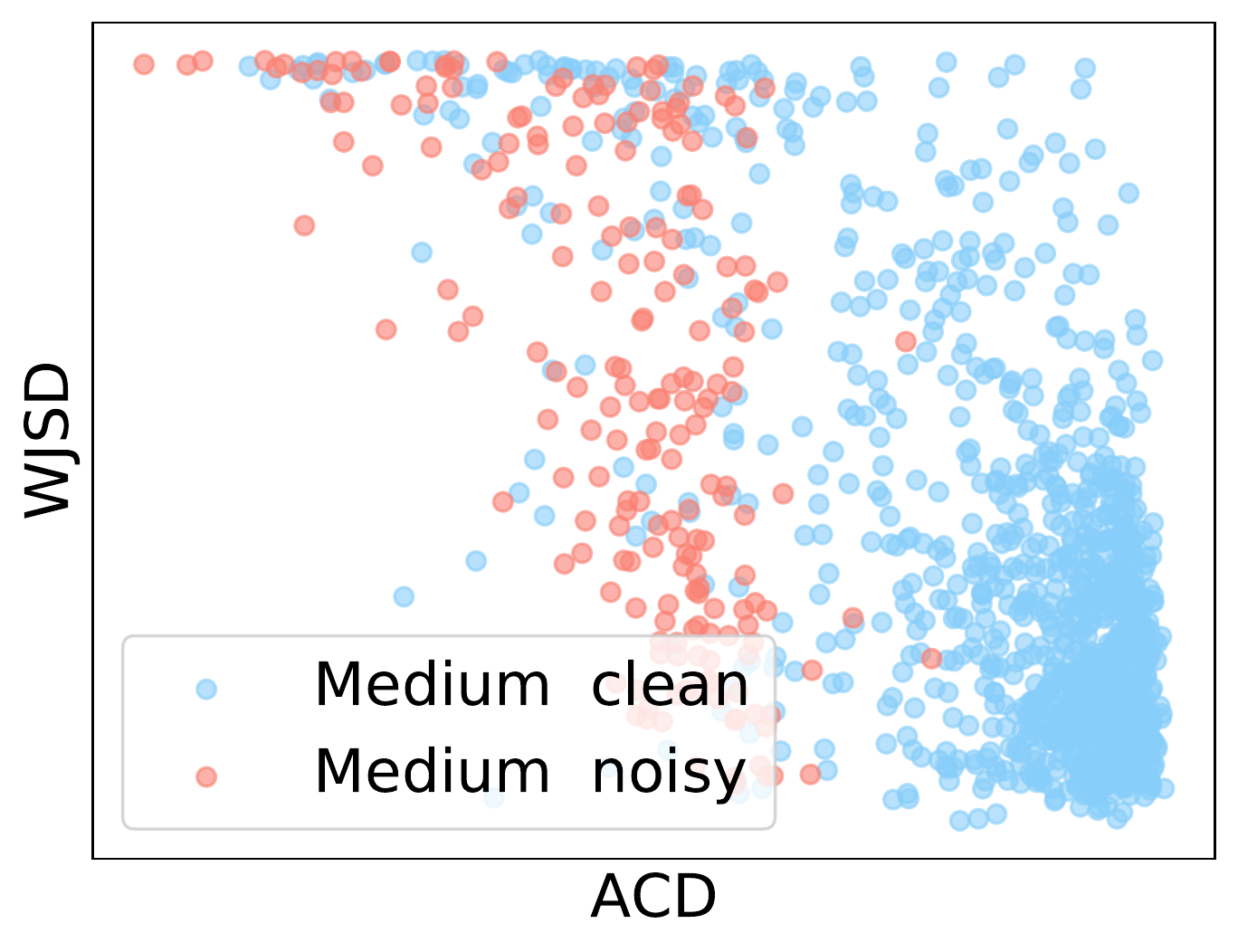}
    \subcaption{Asymmetric noise}
\end{subfigure}
\caption{Scatter plot of the values of the proposed bi-dimensional metrics with (a) symmetric noise and (b) asymmetric noise.}
\label{bi-dimensional_metric_2}
\end{figure}

In this section, we plot the values of bi-dimensional metrics of both clean and noisy samples for head and medium classes under different noise types in \cref{bi-dimensional_metric_2}.
For the case of symmetric noise shown in \cref{bi-dimensional_metric_2} (a), it can be observed that both WJSD and ACD can well separate clean and noisy samples in the head class, while ACD cannot effectively distinguish clean samples from noisy samples in the medium class.
In this case, WJSD shows its advantage in sample separation.
For the case of asymmetric noise shown in \cref{bi-dimensional_metric_2} (b), it can be observed that WJSD cannot distinguish clean samples from noisy samples well for head and medium classes, while ACD can distinguish them well.
It further validates the complementarity of the bi-dimensional metrics.
\section{Hyperparameter Sensitivity Analysis}
\begin{figure}[!h]
\centering
    \includegraphics[width=0.4\textwidth]{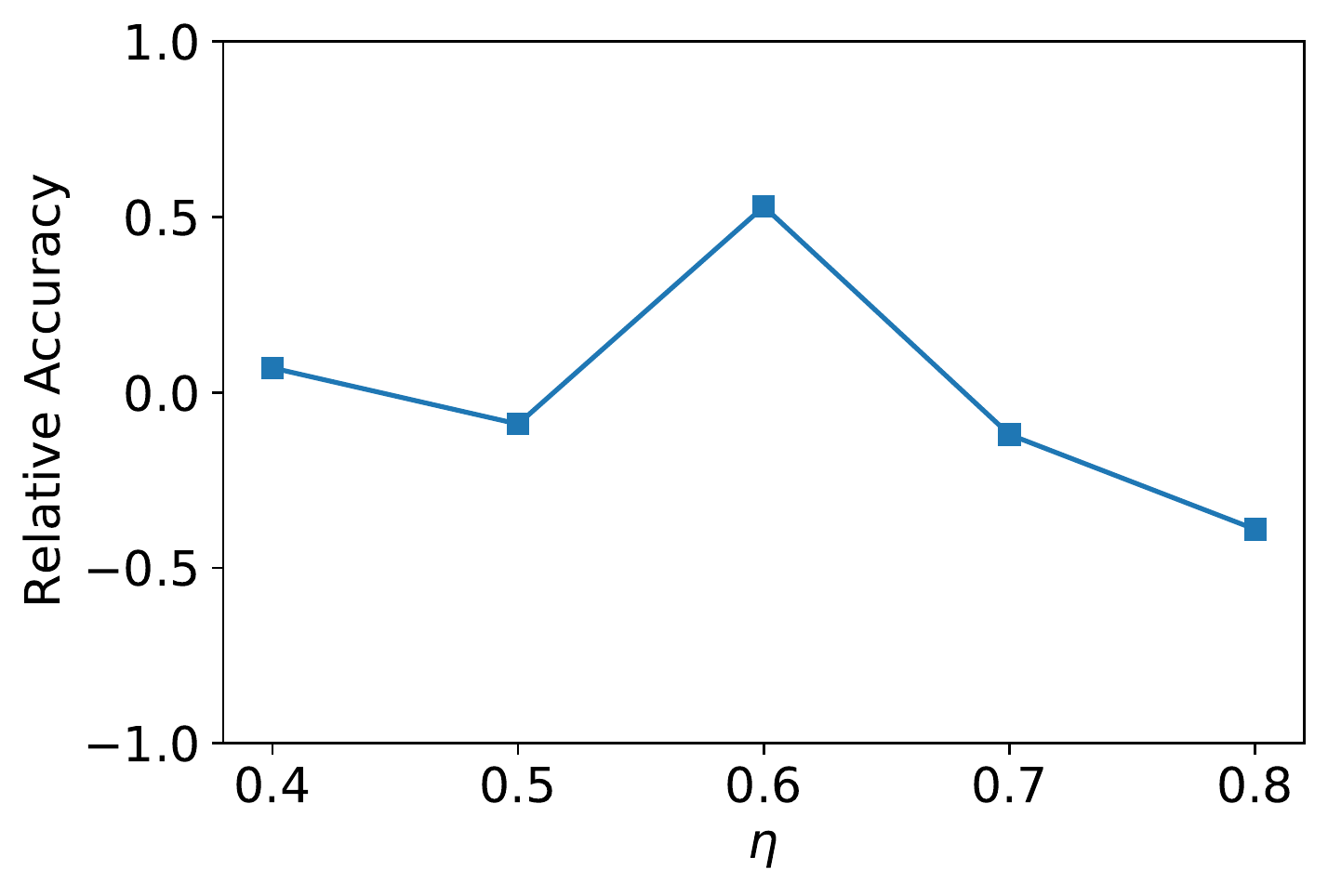}
    \vspace{-5px}
\caption{Impact of hyperparameter $\eta$.}
\label{Hyperparameter}
\end{figure}

We investigate the impact of $\eta$ for dimension selection.
The results are shown in \cref{Hyperparameter}.
We vary $\eta$ from 0.4 to 0.8, and the relative accuracy varies between -0.5\% and 0.5\%.
When $\eta$ is relatively large, more cases not suitable for WJSD separation are WJSD separated, thus compromising the sample separation effect and leading to a decrease in model performance.
Similarly, when $\eta$ is relatively small, cases suitable for WJSD separation are filtered out, affecting model performance.
In general, there is little variation in the accuracy of the model, so the dimension selection we proposed is robust for $\eta$.
\vspace{-10px}
\section{Computational Cost Analysis}
We  compute the relative training cost of each method compared to the baseline on CIFAR-100, as presented in the \cref{Computational cost}.
By comparison, it is evident that the training cost of the proposed method is indeed higher but falls within an acceptable range. Despite this higher cost, the proposed method offers remarkable performance gains in return.
\begin{table}[!h]\footnotesize
        \centering
        \begin{tabular}{@{}lllclll@{}}
                \toprule
             \makecell[c]{Method} &   \makecell[c]{MW-Net} & \makecell[c]{RoLT} & \makecell[c]{HAR} & \makecell[c]{DivideMix}& \makecell[c]{UNICON}& \makecell[c]{Proposed} \\
                \midrule
             {Relative Cost}&      \makecell[c]{$\times$2.60} & \makecell[c]{$\times$2.06} & \makecell[c]{$\times$1.48} & \makecell[c]{$\times$2.28} & \makecell[c]{$\times$2.54} & \makecell[c]{$\times$4.49}\\
          \bottomrule
        \end{tabular}
        \caption{Relative training cost compared to the baseline.}
        \label{Computational cost}
\end{table}
\vspace{-10px}
\section{Experimental Results across Classes}
\begin{figure}[!b]
\vspace{-12px}
\hspace{8px}
\begin{subfigure}[b]{0.5\columnwidth}
\centering
    \includegraphics[width=0.9\textwidth]{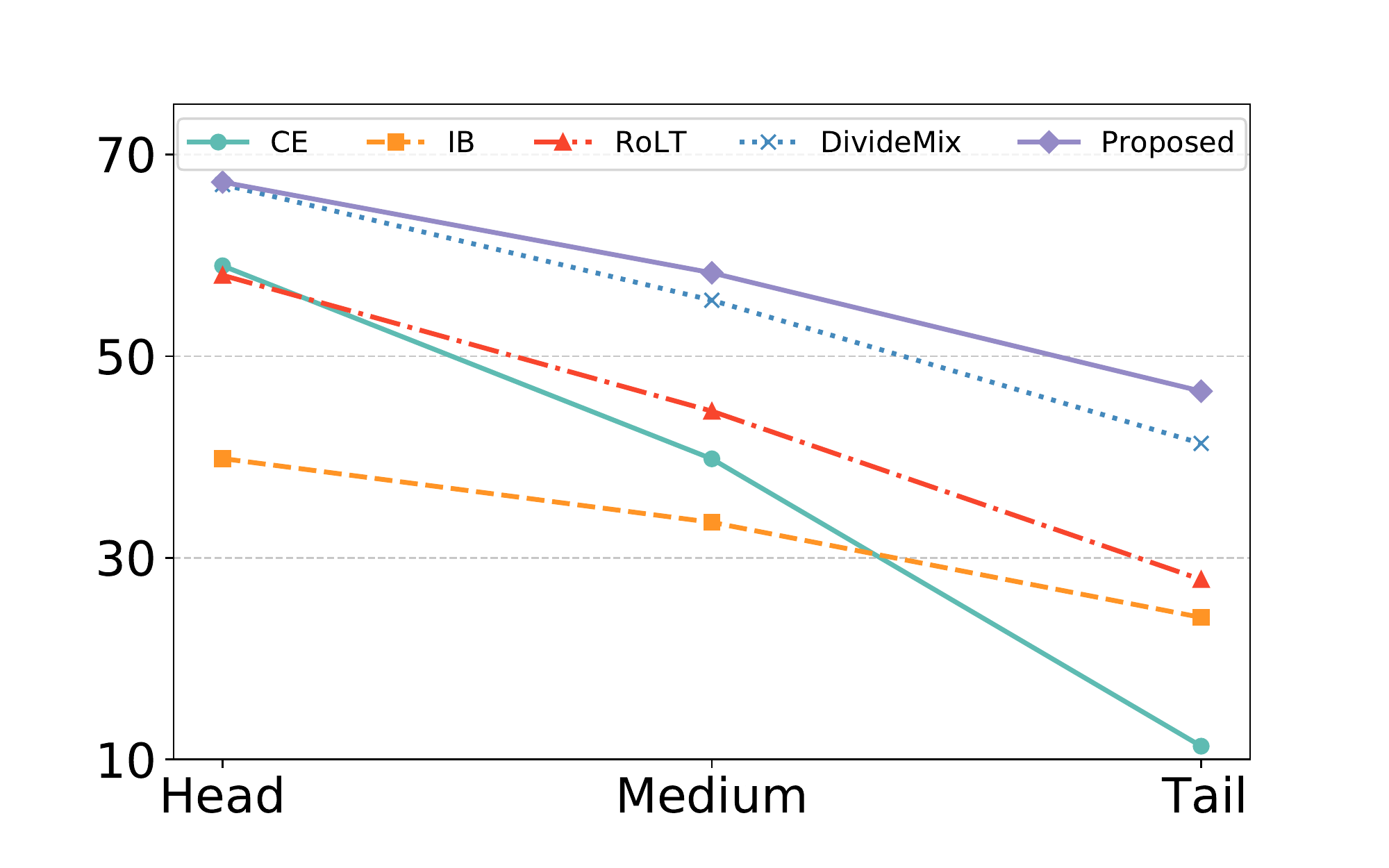}
    \subcaption{Symmetric noise}
\end{subfigure} 
\begin{subfigure}[b]{0.5\columnwidth}
\centering
    \includegraphics[width=0.9\textwidth]{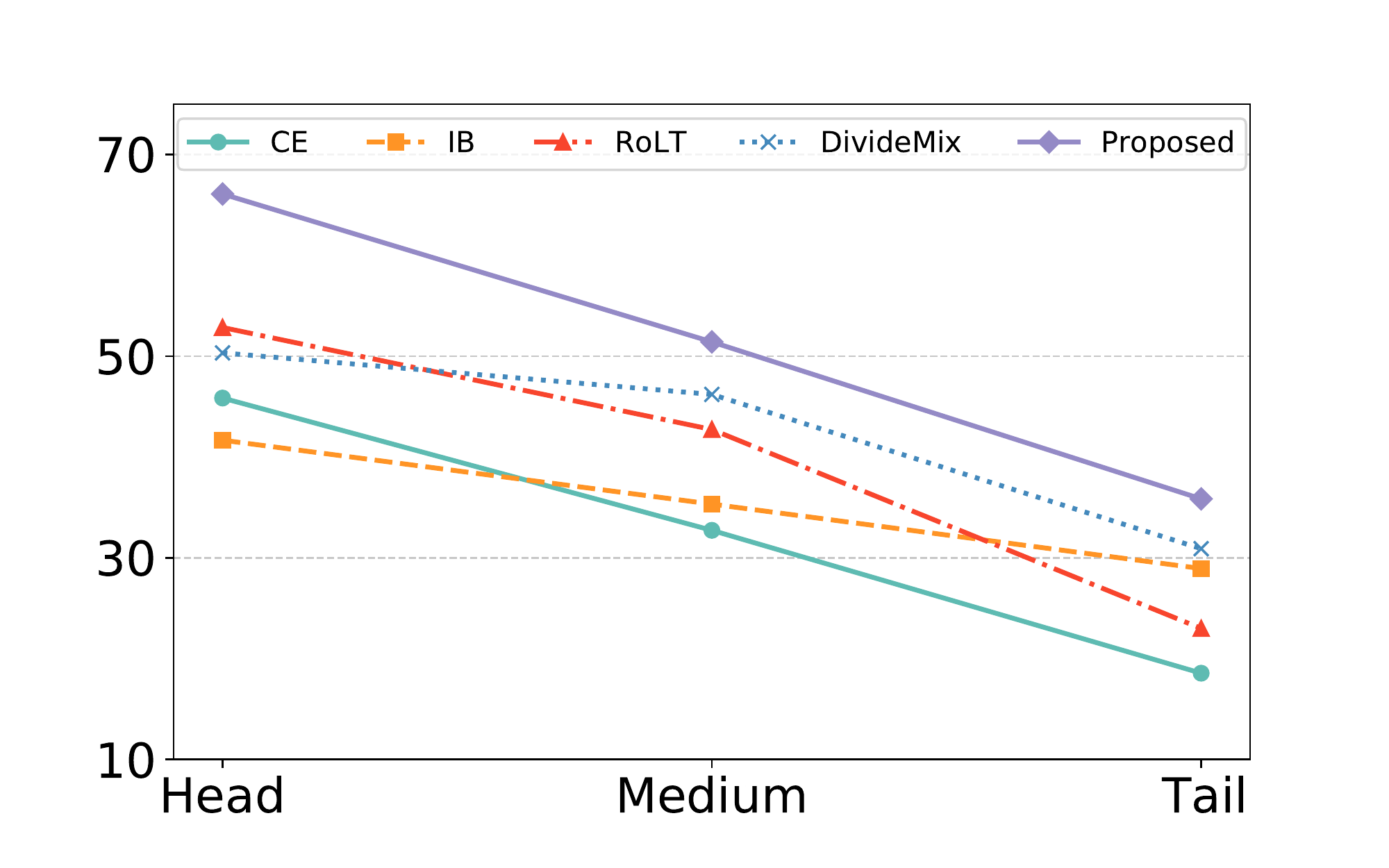}
    \subcaption{Asymmetric noise}
\end{subfigure}
\caption{Line plot of the performance across classes with (a) symmetric noise and (b) asymmetric noise.}
\label{across_classes}
\end{figure}

We conducted a performance comparison of different methods on CIFAR-100 with an imbalance factor of 0.1 and a noise ratio of 0.4 under head, medium and tail classes in \cref{across_classes}.
It can be seen that our approach effectively improves the performance of tail classes without sacrificing the performance of head classes.
The underlying rationale behind is that the proposed WJSD can effectively separate tail classes samples without affecting the separation of  head classes, and ACD are not solely focused on improving the tail classes.

\end{document}